\documentclass[letterpaper]{article} 
\usepackage{aaai25}  
\usepackage{times}  
\usepackage{helvet}  
\usepackage{courier}  
\usepackage[hyphens]{url}  
\usepackage{graphicx} 
\usepackage{natbib}  
\usepackage{caption} 
\usepackage{algorithm}
\usepackage{algorithmic}
\usepackage{color}
\usepackage{amssymb}
\usepackage{amsmath}
\usepackage{pifont}
\usepackage{multirow}
\usepackage{booktabs}
\usepackage{caption}
\usepackage{subcaption}
\usepackage{bm}
\usepackage{newfloat}
\usepackage{listings}
\usepackage[table]{xcolor}
\usepackage[colorlinks=true, citecolor=blue, linkcolor=blue]{hyperref}
\usepackage[capitalise]{cleveref}

\newcommand{\no}[0]{\textcolor{red}{\ding{55}}}
\newcommand{\yes}[0]{\textcolor{green}{\ding{51}}}

\newcommand{\grayrule}[0]{
    \arrayrulecolor{gray!50!white}\midrule\arrayrulecolor{black}
}

\DeclareCaptionStyle{ruled}{labelfont=normalfont,labelsep=colon,strut=off} 
\floatstyle{ruled}
\newfloat{listing}{tb}{lst}{}
\floatname{listing}{Listing}
\title{DepthFM: Fast Generative Monocular Depth Estimation with Flow Matching}
\author{
    Ming Gui\equalcontrib, Johannes Schusterbauer\equalcontrib, Ulrich Prestel, Pingchuan Ma, \\
    Dmytro Kotovenko, Olga Grebenkova, Stefan Andreas Baumann, Tao Hu, Bj\"orn Ommer
}
\affiliations{
    CompVis @ LMU Munich \\
    Munich Center for Machine Learning
}

\begin{document}

\maketitle

\begin{figure*}[!ht]
\centering
\includegraphics[width=0.98\linewidth]{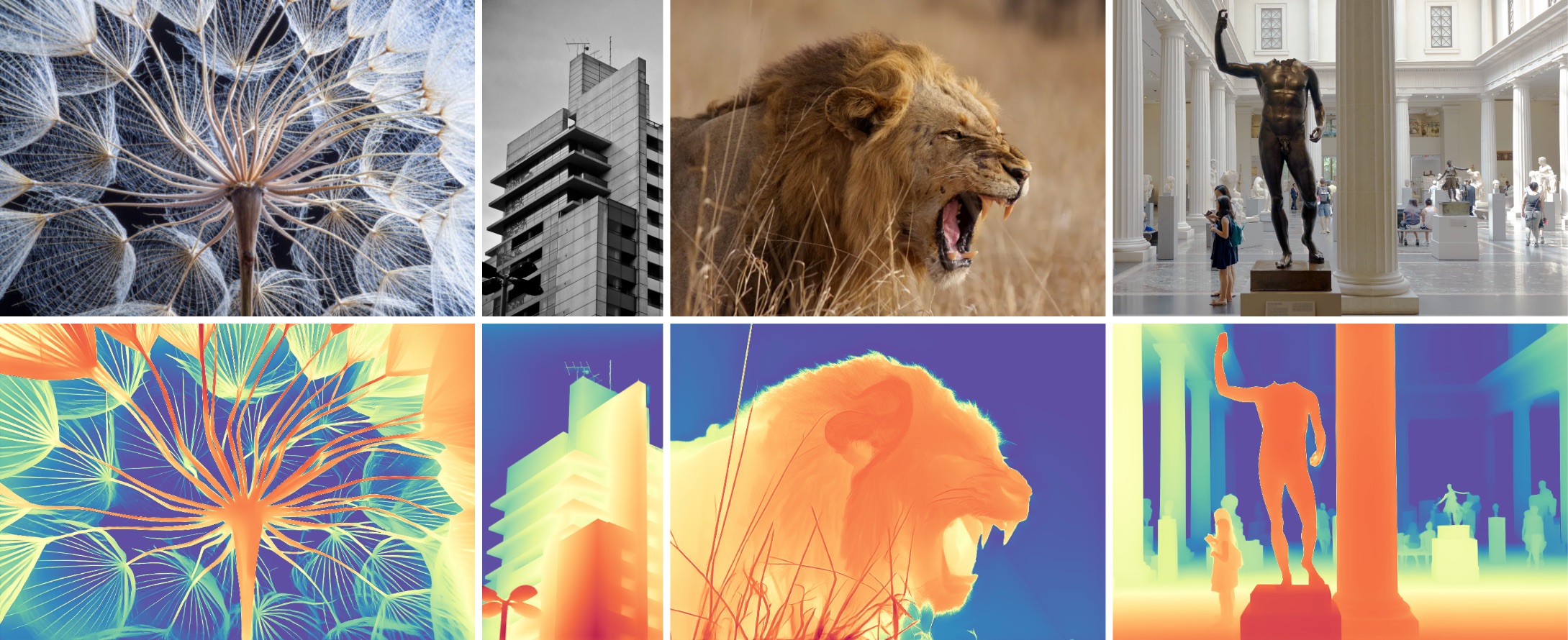}
\caption{We present \textit{DepthFM}, a high-fidelity, fast, and flexible generative monocular depth estimation model.}
\label{fig:teaser}
\end{figure*}

\begin{abstract}
Current discriminative depth estimation methods often produce blurry artifacts, while generative approaches suffer from slow sampling due to curvatures in the noise-to-depth transport. Our method addresses these challenges by framing depth estimation as a direct transport between image and depth distributions. We are the first to explore flow matching in this field, and we demonstrate that its interpolation trajectories enhance both training and sampling efficiency while preserving high performance.
While generative models typically require extensive training data, we mitigate this dependency by integrating external knowledge from a pre-trained image diffusion model, enabling effective transfer even across differing objectives. To further boost our model performance, we employ synthetic data and utilize image-depth pairs generated by a discriminative model on an in-the-wild image dataset. As a generative model, our model can reliably estimate depth confidence, which provides an additional advantage.
Our approach achieves competitive zero-shot performance on standard benchmarks of complex natural scenes while improving sampling efficiency and only requiring minimal synthetic data for training.
\end{abstract}

\begin{links}
    \link{Code}{https://github.com/CompVis/depth-fm}
\end{links}

\section{Introduction}
\label{sec:intro}
Monocular depth estimation is pivotal for 3D scene understanding due to its numerous applications, ranging from core vision tasks such as segmentation \cite{he2021sosd} and visual synthesis \cite{zhang2023adding} to application areas like robotics and autonomous driving \cite{cabon2020vkitti2,kitti}.
Despite the recent strides in this field, estimating realistic geometry from a single image remains challenging. State-of-the-art discriminative depth estimation models exhibit impressive overall performance but still lack fine-grained high-frequency details \cite{yang2024depthanything,yang2024depthanythingv2}. Current generative-based methods \cite{ke2023marigold, fu2024geowizard} address this issue by phrasing depth estimation as an image-conditional iterative denoising process using diffusion models \cite{ho2020denoising, song2021scorebased}. Although they can generate realistic and accurate depth maps, these methods suffer from extremely long inference times because of the integration over a highly curved ordinary differential equation~(ODE) trajectory.

Flow Matching (FM) \cite{lipman2022flow,rectifiedflow_iclr23,albergo2023stochastic,albergo2022building,actionmatching} is an attractive alternative paradigm. These methods emerged as a strong competitor to the currently prominent Diffusion Models~(DM) \cite{sohl2015deep,ho2020denoising,song2021scorebased} and offer enhanced flexibility in trajectory design and starting distribution.
While diffusion models offer samples of great diversity, the curved diffusion trajectories through solution space entail high computational costs. Conversely, the straighter trajectories of flow matching result in much faster processing \cite{lipman2022flow,lee2023minimizing}. We hypothesize that these characteristics of flow matching are a much better fit for image-based depth estimation, in contrast to diffusion models. To further enhance training and inference efficiency, we use data-dependent couplings \cite{schusterbauer2023boosting} for our model, which we call \textit{DepthFM}. Unlike conventional diffusion models that start the generative process from noise and end with a depth map, our method directly models the trajectory from image to depth space.

However, there are several challenges in training efficient generative depth estimation models. First, the computational requirements for training generative models are extremely high~\cite{zhang2024improving}. Second, annotating depth is very difficult~\cite{geiger2012we_kitti}, making data efficiency a critical issue. To address these problems, we propose to seek external knowledge from a pre-trained \textit{image diffusion model} and a pre-trained \textit{discriminative depth estimation model}. 
We incorporate a strong image prior from unsupervised generative training together with a strong depth prior from a pre-trained discriminative model while preserving the advantages inherent to the generative approach. We augment our model with prior information by fine-tuning our approach from an image synthesis foundation model, specifically, SD2.1 \cite{rombach2022high}. We show the feasibility of transferring information between DM and FM by fine-tuning a flow matching model from a diffusion model prior. This provides our model with initial visual cues and significantly speeds up training. It also allows us to train exclusively on a small amount of \emph{synthetic data} and still achieve robust generalization to real-world images.

In summary, to improve the sampling, training, and data efficiency in generative depth estimation, our contributions are as follows: 

\begin{itemize}
    \item We introduce DepthFM, a versatile and fast generative model for monocular depth estimation. We are the first to formulate monocular depth estimation as a direct transport problem, represented via flow matching. By utilizing more flexible trajectory and distribution choices compared to diffusion models, DepthFM enhances sampling efficiency, leading to more efficient depth estimation. 
    \item To enhance training and data efficiency, we leverage external knowledge from both the generative and discriminative communities. We successfully transfer a robust image prior from a pre-trained diffusion model to a flow matching model, significantly reducing reliance on training data. Additionally, we demonstrate that a pre-trained, off-the-shelf discriminative model can further boost the performance of generative depth estimation.
    \item Ultimately, our findings show that flow matching is highly efficient and capable of synthesizing depth maps in a single inference step. Despite being trained solely on synthetic data, DepthFM delivers competitive or state-of-the-art performance on benchmark datasets and natural images. Additionally, our method demonstrates state-of-the-art performance in depth completion.
\end{itemize}

\section{Related Works}
\label{sec:rel}

Depth estimation can be broadly divided into discriminative and generative methods. Prominent examples of discriminative methods include MiDaS~\cite{Ranftl2020MiDaS}, Depth Anything~\cite{yang2024depthanything,yang2024depthanythingv2}, and Metric3D~\cite{yin2023metric3d,hu2024metric3d}.

Recently, generative diffusion-based models have been explored for affine-invariant depth estimation \cite{ke2023marigold, fu2024geowizard}. These models exploit the rich knowledge embedded in large-scale vision foundation models to produce high-quality depth maps. Despite their promising results, diffusion-based methods face significant challenges. First, they suffer from slow sampling times, even when using ODE approximations to solve the underlying SDEs. Second, the diffusion formulation relies on a Gaussian source distribution~\cite{sohl2015deep,song2019generative}, which may not fully capture the natural relationship between images and their corresponding depth maps.

In contrast, flow matching-based models \cite{lipman2022flow,rectifiedflow_iclr23,albergo2023stochastic} have demonstrated promise across various tasks and offer faster sampling speeds~\cite{rectifiedflow_iclr23}. Additionally, optimal transport can be achieved even when source distribution deviates from a Gaussian distribution, which is advantageous for certain tasks \cite{tong2023improving}. Our work takes a first step in using flow matching for monocular depth estimation, aiming to reduce sampling costs by leveraging the inherent straight sampling trajectory~\cite{lee2023minimizing}. Furthermore, we seek to combine the strengths of both discriminative and generative depth estimation by designing a simple knowledge distillation pipeline and transferring knowledge from discriminative models to enhance generative depth estimation.

To the best of our knowledge, no prior work has explored the use of flow matching to facilitate the distribution transfer between images and depth maps, despite their intuitive proximity compared to noise and depth maps. In addition, we aim to improve training efficiency by incorporating insights from diffusion models, exploiting the similarity of their objectives~\cite{lee2023minimizing}. We discuss additional related work in the appendix.

\begin{figure}[t]
    \centering
        \includegraphics[width=\linewidth]{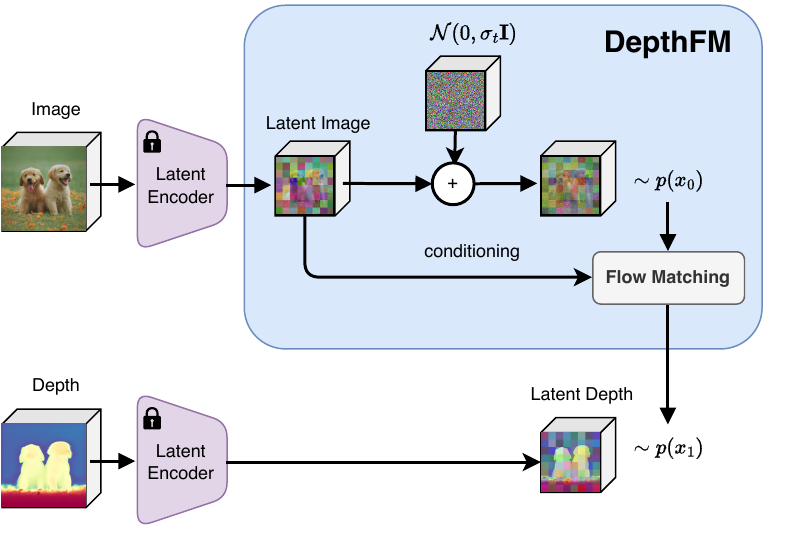}
    \caption{
    Overview of our training pipeline. We use flow matching to regress the vector field between the image latent $x_0$ and the corresponding depth latent $x_1$.
    }
    \label{fig:pipeline}
\end{figure}

\section{Methodology}
\label{sec:method}

\subsection{Background: Flow Matching}
\label{sec:flow_matching}
Flow Matching~\cite{lipman2022flow,rectifiedflow_iclr23,albergo2023stochastic,actionmatching} belongs to the category of generative models designed to regress vector fields based on fixed conditional probability paths. Denote $\mathbb{R}^d$ as the data space with data points $x$ and $u_t(x): [0,1] \times \mathbb{R}^d \rightarrow \mathbb{R}^d$ the time-dependent vector field, which defines the ODE $dx=u_t(x)dt$. Let $\phi_t(x)$ represent the solution to this ODE with the initial condition $\phi_0(x)=x$.

The probability density path $p_t:[0,1]\times \mathbb{R}^d \rightarrow \mathbb{R}_{>0}$ characterizes the probability distribution of $x$ at timestep $t$ with $\int p_t(x)dx=1$. According to the continuity condition, the pushforward function $p_t=[\phi_t]_{\#}(p_0)$ then transports the probability density path $p$ along $u$ from timestep $0$ to $t$.

~\citet{lipman2022flow} showed that we can efficiently train a neural network using the conditional flow matching objective, to regress conditional vector fields $u_t(x|x_1)$ by sampling $p_t(x|x_1)$:
\begin{equation}
    \mathcal{L}_{FM}(\theta)=\mathbb{E}_{t,p_{data}(x_1),x\sim p_t(x|x_1)}||v_\theta(t,x)-u_t(x|x_1)||.
\end{equation}
The most common approach to define $u_t(x|x_1)$ is as a straight path from $x_0$ to $x_1$~\cite{rectifiedflow_iclr23,lipman2022flow}.
Assuming that $p_0$ is a standard Gaussian with $x_0 = \epsilon \sim \mathcal{N}(0, \mathbb{I})$, the intermediate interpolant is defined as $x_t=tx_1 + (1-t)\epsilon$, and a valid vector field that satisfies the probability path is given by $u_t(x|x_1)=\frac{x_1 - x}{1-t}$.

\subsection{Data Coupling in FM for Depth Estimation} 
\label{sec:data_dependent}

The monocular depth estimation task involves training a function that uses image conditioning to generate depth. Let us denote the image in pixel space $X_0$ and the corresponding depth in pixel space $X_1$. 

\subsubsection{Flow in Latent Space}
In order to reduce the computational demands associated with training FM models for high-resolution depth estimation synthesis, we follow \cite{rombach2022high, dao2023flow_lfm, schusterbauer2023boosting,hulfm, ke2023marigold} and utilize an autoencoder model that provides a compressed latent space that aligns perceptually with the image pixel space. This approach also facilitates the direct inheritance of a robust model prior obtained from foundational LDMs such as Stable Diffusion.
Similar to \cite{ke2023marigold} we move both modalities (i.e., RGB images and depths) to the latent space. Without further mentioning, $x_0=\texttt{Enc}(X_0)$, $x_1=\texttt{Enc}(X_1)$ take place in the latent space. We can accordingly use the {Decoder} (\texttt{Dec}) to decode them back to image and depth space. Note that we employ a different normalization strategy compared to Marigold \cite{ke2023marigold} in that we use log-scaled depth which ensures a more balanced capacity allocation for both indoor and outdoor scenes. Further details, including explanations and ablation studies, can be found in the appendix.

\subsubsection{Direct Transport between Image and Depth} Let \((x_0, x_1)\) be ground truth image-depth feature pairs. In contrast to diffusion-based depth estimators~\cite{ke2023marigold, fu2024geowizard}, which map Gaussian noise \(\epsilon\) with image conditioning to depth \(x_1\) by $p(x_1|\epsilon;x_0)$, we formulate depth estimation as a direct distribution transport between the image feature \(x_0\) and the depth feature \(x_1\) by $p(x_1|x_0)$, which can be effectively solved using conditional flow matching~\cite{tong2023improving_alex,albergo2023stochastic}. Specifically, we model the intrinsic transport relation between the image feature \(x_0\) and the depth feature \(x_1\).
As shown in~\Cref{tab:motivate_coupling}, using paired coupling between image and depth maps results in a far shorter transport path than mapping directly from Gaussian noise to depth maps. 
The transport between the image and the depth distribution can be defined as
\begin{equation}
    x_t \sim p_t(x|(x_0,x_1))=\mathcal{N}(x|tx_1+(1-t)x_0,\sigma_{\min}^2\mathbb{I}),
\end{equation}
\begin{equation}
    u_t(x|(x_0,x_1))=x_1 - x_0,
\end{equation}
where $u_t(x|(x_0,x_1))$ is the vector field that transports $x_0$ along space to the marginal distribution $p_t(x|(x_0,x_1))$. To avoid singularity problems, we smooth both $p(x_0)$ and $p(x_1)$ with a minimum smoothing factor $\sigma_{\min}$ to obtain the corresponding data distributions $\mathcal{N}(x_0, \sigma_{\min}^2)$ and $\mathcal{N}(x_1, \sigma_{\min}^2)$. 
 
Despite the different modalities and data manifolds of $x_0$ and $x_1$, the optimal transport condition between $p(x_0)$ and $p(x_1)$ is inherently satisfied due to image-to-depth pairs. This addresses the dynamic optimal transport problem in the transition for image-to-depth translation within the flow matching paradigm, ensuring more stable and faster training \cite{tong2023improving}. The loss thus takes the form:
\begin{equation}
    \mathcal{L}(\theta)= \mathbb{E}_{t\sim\mathcal{U}[0,1],(x_0,x_1)\sim \mathcal{D}^{GT}}||v_\theta(t,x_t;\bar{x})- (x_1-x_0)||.
\end{equation}
Here, \(\bar{x}\) represents a clean copy of the image latent feature \(x_0\) and serves as additional conditioning, $\mathcal{D}^\text{GT}$ refers to an image depth dataset. Notably, we improve the transport by injecting the image feature signal into the model, formulating it as \(v_\theta(t,x_t;\bar{x})\) instead of \(v_\theta(t,x_t)\).

\begin{table}[]
    \centering
    \footnotesize
    \begin{tabular}{lcc}
    \toprule
     Coupling Pattern & EMD-$L_{2}$ $\downarrow$ & EMD-$L_{1}$ $\downarrow$\\
    \midrule
        Random  &  0.981 & 0.974 \\
        Paired (\textit{Ours}) &  \textbf{0.686} & \textbf{0.691} \\
    \bottomrule
    \end{tabular}
    \caption{Paired (Image to Depth) Coupling is better than the Random Coupling. EMD = Earth Mover's Distance.}
    \label{tab:motivate_coupling}
\end{table}

\subsubsection{Noise Augmentation}
Noise augmentation, first used in cascaded diffusion models~\cite{ho2022cascaded}, improves the performance of super-resolution models by adding Gaussian noise to the low-resolution conditioning signal. We extend this technique and apply Gaussian noise augmentation to the terminal distribution at $t=0$. We define the terminal data points $x_0$ as a weighted combination between the image feature $\bar{x}$ and the Gaussian noise~$\epsilon$: $ {x}_{0} := \sqrt{\bar{\alpha}_{t_s}} \bar{x} + \sqrt{1-\bar{\alpha}_{t_s}}  \epsilon$,
where $\bar{\alpha}_{t_s} \in [0,1]$ can be any predefined variance-preserving noise schedule~\cite{ho2020denoising,song2021scorebased, nichol2021improved}, and $t_s$ is a hyperparameter. We choose to augment the feature in a way that preserves the variance of the data point. We hypothesize that adding a small amount of Gaussian noise will smooth the base probability density and keep it well-defined over a larger manifold. In addition, adding this stochasticity allows us to use it in the sampling process for further applications, such as indicating confidence.
In particular, unlike previous work on noise augmentation using noise-image pairs, we empirically show its feasibility on image-depth pairs, which has not been explored before.

\subsection{Dual Knowledge Transfer for better Training and Data Efficiency}
Training depth estimation models faces significant challenges, including high computational demands during training \cite{zhang2024improving} and a lack of high-quality annotated depth data \cite{geiger2012we_kitti}. To address these issues, we propose to leverage external knowledge from pre-trained image diffusion models and pre-trained discriminative depth estimation models and combine their strengths to improve training efficiency and performance.

\subsubsection{Image Prior for Training Efficiency}
Intuitively, a visual synthesis model that generates sound images must also have some notion of the inherent depth of a scene. Similar to \cite{ke2023marigold}, we use a large pre-trained generative image model that has been trained on a large amount of data, and use this prior knowledge to infer depth. Diffusion models can be trained with different parameterizations, including the $x$, $\epsilon$, and ${v}$ parameterizations. A model parametrized with $v$ regresses the \textit{velocity} between samples from the two terminal distributions \cite{salimans2022progressive}.
In the context of FM, where the two terminal distribution samples are denoted as $x_0$ and $x_1$, the objective of the $v$ parameterization can be mathematically formulated as ${v}=\alpha_t x_0 - \sigma_t x_1$, where $\alpha_t$ and $\sigma_t$ is the fixed diffusion schedule. In comparison, our FM objective regresses a vector field of $\mathbf{v}=x_1-x_0$.
The similarity between the DM and FM objectives allows us to fine-tune the pre-trained diffusion model with the conditional flow matching loss and use the strong image prior to speed up convergence for generative depth estimation.

\subsubsection{Depth Prior for Data Efficiency}
Generative depth estimation models \citep{ke2023marigold, fu2024geowizard} offer high visual fidelity but often lag behind discriminative models in quantitative metrics \citep{hu2024metric3d, yang2024depthanything}. While the high performance of discriminative methods is often due to training on a large amount of data, the higher fidelity of generative models is due to the ability to sample from the conditional posterior instead of the mean prediction of discriminative counterparts.
To be data efficient while still producing high-fidelity depth maps, we integrate the strengths of both model types. Using a discriminative depth model as a teacher, we improve our generative model, increasing both robustness and performance.

In detail, let us denote the discriminative monocular depth estimation teacher model as $T$. We use this model to predict depth on an unlabeled in-the-wild image dataset $\hat{\mathcal{D}}^u$ and generate discriminative samples ${\mathcal{D}}^u$ in the form of ${\mathcal{D}}^u=\{(u_i, T(u_i)) | u_i \in \hat{\mathcal{D}}^u\}$. The loss $\mathcal{L}$ can be formulated as:
\begin{equation}
    \mathcal{L}(\theta)= \mathbb{E}_{t\sim\mathcal{U}[0,1],(x_0,x_1)\sim \mathcal{D}}||v_\theta((t,x_t);\bar{x})- (x_1-x_0)||,
    \label{eq:mix_dataset}
\end{equation}
where $\mathcal{D} = \mathcal{D}^\text{GT} \cup  \texttt{part}(\mathcal{D}^{u},k) $ and $\texttt{part}()$ is the partition function that randomly selects samples from $\mathcal{D}^{u}$ such that the number of selected samples corresponds to $|\texttt{part}(\mathcal{D}^{u},k)| = k \cdot |\mathcal{D}^\text{GT}|$.

\section{Experiments}
\label{sec:exp}

\subsection{Training and Evaluation Details}
We train our depth estimation model on two synthetic datasets, Hypersim \cite{roberts2021hypersim} and Virtual KITTI \cite{cabon2020vkitti2} to cover both indoor and outdoor scenes. Following \cite{ke2023marigold} we take 54K training samples from Hypersim and 20K training samples from Virtual KITTI.
We leverage Metric3D v2 \cite{hu2024metric3d}, as our teacher model. We apply this model on a subset of the general-purpose image dataset \textit{Unsplash} \cite{unsplash}, to generate image-depth pairs.

We perform zero-shot evaluations on established real-world depth estimation benchmarks NYUv2~\cite{silberman12nyuv2}, KITTI~\cite{behley2019semantickitti}, ETH3D~\cite{schops2017multiEth3d}, ScanNet~\cite{dai2017scannet}, and DIODE~\cite{diode_dataset}. Let $d$ be the ground truth depth and $\hat{d}$ be the predicted depth. The evaluation metrics are the Absolute Mean Relative Error (\textbf{RelAbs}), calculated as $\frac{1}{M} \sum_{i=1}^M |d_i - \hat{d_i}|/d_i$ and $\bm{\delta1}$  \textbf{accuracy}, which measures the ratio of all pixels satisfying $\max(d_i/\hat{d_i}, \hat{d_i}/d_i) < 1.25$. Unless otherwise specified, we evaluate our model using an ensemble size of $10$ and $4$ Euler steps, and scale and shift our predictions to match the ground truth depth in log space.

\begin{table*}[t]
\setlength\tabcolsep{2.5pt}
\centering
\footnotesize
\begin{tabular}{clrrcccccccccc}
\toprule
\multirow{2}{*}{} &
\multicolumn{1}{l}{\multirow{2}{*}{\textbf{Method}}} &
\multicolumn{2}{c}{\textbf{\#Train samples}} &
\multicolumn{2}{c}{\textbf{NYUv2}} &
\multicolumn{2}{c}{\textbf{KITTI}} &
\multicolumn{2}{c}{\textbf{ETH3D}} &
\multicolumn{2}{c}{\textbf{ScanNet}} &
\multicolumn{2}{c}{\textbf{DIODE}}
\\
\cmidrule(lr){3-14} & \multicolumn{1}{l}{} & \multicolumn{1}{c}{Real} & \multicolumn{1}{c}{Synthetic} & AbsRel$\downarrow$ & $\delta$1$\uparrow$ & AbsRel$\downarrow$ & $\delta$1$\uparrow$ & AbsRel$\downarrow$ & $\delta$1$\uparrow$ & AbsRel$\downarrow$ & $\delta$1$\uparrow$ & AbsRel$\downarrow$ & $\delta$1$\uparrow$
\\
\midrule
\multirow{7}{*}{\rotatebox{90}{\small\textit{Discriminative}}}
    & MiDaS \color{gray}\tiny\cite{Ranftl2020MiDaS} & 2M & ---
        & 0.111 & 88.5 & 0.236 & 63.0 & 0.184 & 75.2 & 0.121 & 84.6 & 0.332 & 71.5 \\
    & Omnidata \color{gray}\tiny\cite{eftekhar2021omnidata} & 11.9M & 301K
        & 0.074 & 94.5 & 0.149 & 83.5 & 0.166 & 77.8 & 0.075 & 93.6 & 0.339 & 74.2 \\
    & HDN \color{gray}\tiny\cite{zhang2022hierarchical} & 300K & ---
        & 0.069 & 94.8 & 0.115 & 86.7 & 0.121 & 83.3 & 0.080 & 93.9 & 0.246 & 78.0 \\
    & DPT \color{gray}\tiny\cite{Ranftl_2021_ICCV_DPT} & 1.2M & 188K
        & 0.098 & 90.3 & 0.100 & 90.1 & 0.078 & 94.6 & 0.082 & 93.4 & 0.182 & 75.8 \\
    & DA \color{gray}\tiny\cite{yang2024depthanything} & 1.5M & 62M
        & 0.043 & 98.1 & 0.076 & 94.7 & 0.127 & 88.2 & --- & --- & 0.066 & 95.2 \\
    & DAv2 \color{gray}\tiny\cite{yang2024depthanythingv2} & --- & 595K+62M
        & 0.044 & 97.9 & 0.075 & 94.8 & 0.132 & 86.2 & --- & --- & 0.065 & 95.4 \\
    & Metric3D v2 \color{gray}\tiny\cite{hu2024metric3d} & 25M & 91K
        & 0.043 & 98.1 & 0.044 & 98.2 & 0.042 & 98.3 & 0.022$^\dagger$ & 99.4$^\dagger$ & 0.136 & 89.5 \\
\midrule
    \multirow{4}{*}{\rotatebox{90}{\small\textit{Generative}}}
    & Marigold \color{gray}\tiny\cite{ke2023marigold} & --- & 74K
        & 0.055 & 96.4 & 0.099 & 91.6 & 0.065 & 96.0 & 0.064 & 95.1 & 0.308 & 77.3 \\
    & GeoWizard \color{gray}\tiny\cite{fu2024geowizard} & --- & 280K
        & 0.052 & 96.6 & 0.097 & 92.1 & 0.064 & 96.1 & 0.061 & 95.3 & 0.297 & 79.2 \\ 
    \arrayrulecolor{gray!50!white}\cmidrule(l){2-14}\arrayrulecolor{black}
    & DepthFM-I & --- & 74K
        & 0.060 & 95.5 & 0.091 & 90.2 & 0.065 & 95.4 & 0.066 & 94.9 & 0.224 & 78.5 \\
    & DepthFM-ID & --- & 74K+7.4K
        & 0.055 & 96.3 & 0.089 & 91.3 & 0.058 & 96.2 & 0.063 & 95.4 & 0.212 & 80.0 \\
\bottomrule
\end{tabular}
\caption{Quantitative comparison with affine-invariant depth estimators on \textit{zero-shot} benchmarks. $\delta 1$ is presented in percentage. Our method shows competitive performance across datasets. DepthFM-I and DepthFM-ID refer to our model trained with image prior and image-depth prior, respectively. DA stands for the Depth Anything model family. Some baselines are sourced from Marigold \citep{ke2023marigold} and GeoWizard \citep{fu2024geowizard}. State-of-the-art discriminative models, which heavily rely on \textit{extensive} amounts of annotated training data, are listed in the upper part of the table. $\dagger$: Models are trained with normals.
}
\label{tab:sota}
\end{table*}

\begin{figure}
\center \small
    \setlength\tabcolsep{1pt}

    \newcommand{\imagepng}[2]{
    \includegraphics[width=0.22\linewidth]{figs/pred-over-nfes/#1/#2.png}
    }
    
    \begin{tabular}{cccc}
        \imagepng{im3}{original}
        & \imagepng{im3}{marigold_nfe2}
            & \imagepng{im3}{our_nfe2}
            & \rotatebox{270}{\hspace{-7.5em} NFE $=2$} \\
        & \imagepng{im3}{marigold_nfe4}
            & \imagepng{im3}{our_nfe4}
            & \rotatebox{270}{\hspace{-7.5em} NFE $=4$} \\
        & Marigold & DepthFM \tiny(\textit{Ours}) \\
    \end{tabular}
    \caption{Zero-shot qualitative comparison with few inference steps. DepthFM can output realistic depth maps with just two inference steps.
    }
    \label{fig:pred-over-nfes}
\end{figure}

\begin{figure}
    \centering
    \includegraphics[width=0.85\linewidth]{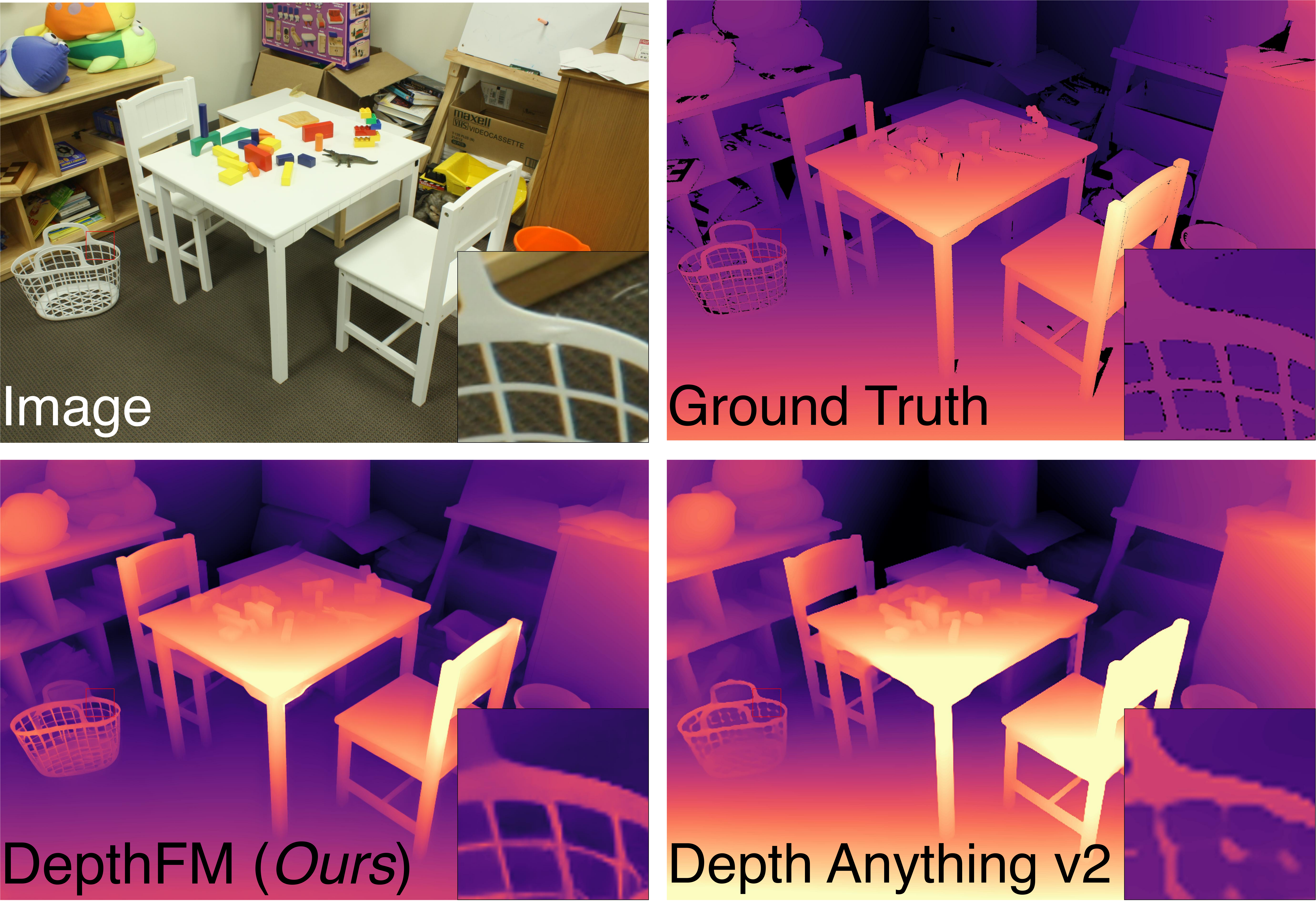}
    \caption{Depth predictions from high-resolution RGBD Middlebury-2014 dataset. Best viewed when zoomed in.}
    \label{fig:fidelity-zoom-ins}
\end{figure}

\begin{table}[t]
    \centering
    \footnotesize
    \begin{tabular}{lcccc}
        \toprule
        NFEs          & 1     & 2     & 4     & 10    \\ \midrule
        Marigold      & 48.8 & 71.5 & 82.7 & 94.8 \\
        DepthFM \tiny(\textit{Ours}) & \textbf{95.0} & \textbf{95.6} & \textbf{96.3} & \textbf{96.2} \\
        \bottomrule
    \end{tabular}
    \caption{$\delta 1$ evaluation on NYUv2 for different numbers of function evaluations (NFE). We fix the ensemble size to $10$.
    }
    \label{tab:fm-marigold-nfe}
\end{table}

\begin{figure*}[t]
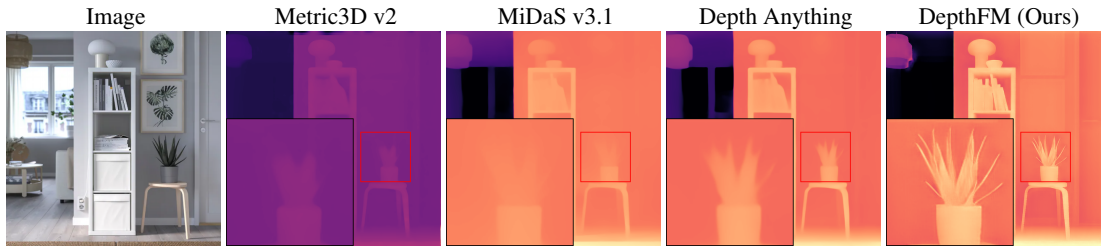

    \setlength\tabcolsep{0.01pt}
    \centering
    \small
    \newcommand{\imgwidth}{0.16\textwidth}
    \newcommand{\imagepng}[2]{
    \includegraphics[width=\imgwidth]{figs/zoom_ins/#1/#1_#2.png}
    }
    \newcommand{\rowpng}[1]{
        \imagepng{#1}{original} &
        \imagepng{#1}{metric3d} &
        \imagepng{#1}{midas} &
        \imagepng{#1}{da} &
        \imagepng{#1}{ours}
    }
    \begin{tabular}{ccccc}
        Image &
        Metric3D v2 &
        MiDaS v3.1  &
        Depth Anything &
        {DepthFM (Ours)}
        \\
        \rowpng{ikea} \\
    \end{tabular}
    \caption{Qualitative comparison of our method against discriminative methods. Best viewed when zoomed in.}
    \label{fig:disc-qualitative}
\end{figure*}

\subsection{Zero-shot Depth Estimation}

\paragraph{Main Quantitative Result}
\Cref{tab:sota} compares our model quantitatively with state-of-the-art depth estimation methods. Unlike other approaches that often require large training datasets, our method leverages the rich knowledge of a diffusion-based foundation model (\textit{image prior}) and a discriminative teacher model (\textit{depth prior}). This strategy reduces the computational burden while emphasizing our approach's adaptability and training efficiency. By training only on 74k synthetic samples and an additional 7.4k samples from a discriminative depth estimation method, our model demonstrates exceptional generalization and achieves high \textit{zero-shot} depth estimation performance on both indoor and outdoor datasets.

\paragraph{Comparison against Generative Models}
Our DepthFM model achieves remarkable sampling efficiency without sacrificing performance. To highlight its inference speed, we quantitatively compare it to Marigold~\cite{ke2023marigold}, a representative diffusion-based generative model. Both models share the same foundational image synthesis model~(SD2.1) and network architecture and only differ in the respective training objective and starting distribution (noise versus image). \Cref{tab:fm-marigold-nfe} shows that DepthFM consistently outperforms Marigold in the low number of function evaluations (NFE) regime, achieving superior results even with \textit{one} NFE, compared to Marigold's performance with \textit{four} NFEs. This is further supported qualitatively in \Cref{fig:pred-over-nfes}, where DepthFM delivers high-quality results with minimal sampling steps, while Marigold requires more steps to produce reasonable results. These results demonstrate the efficiency and effectiveness of DepthFM for fast, high-quality depth estimation.

\paragraph{Comparison against Discriminative Models}
Despite recent advances in discriminative depth estimation, blurred lines at the edges of objects remain a common problem, resulting in lower fidelity. Generative methods overcome this problem and produce sharper depth predictions by allowing sampling from the image-conditional posterior distribution of valid depth maps. In contrast, discriminative models are prone to mode averaging and lack detail, as illustrated in ~\Cref{fig:fidelity-zoom-ins} \& \Cref{fig:disc-qualitative}. To further quantify fidelity, we use the method introduced by \cite{Hu2018RevisitingSI}. On the high-resolution Middlebury-2014 dataset \cite{scharstein2014high}, we predict depth maps, extract edges using a Sobel filter, and then measure edge precision and recall. Higher precision indicates sharper and more precise edges, while higher recall reflects the accuracy of the predicted edges. \Cref{tab:precision-recall} shows that our method significantly outperforms state-of-the-art discriminative depth estimation models in terms of fidelity, achieving higher precision and recall. We further explain the metrics and additional qualitative results in the appendix.

\begin{figure}
    \center \small
    \setlength\tabcolsep{1pt}

    \newcommand{\imagepng}[1]{
    \includegraphics[width=0.3\linewidth]{figs/uncertainty/#1_dog-512.png}
    }
    
    \begin{tabular}{ccc}
        Image & Depth Prediction & Uncertainty \\
        \imagepng{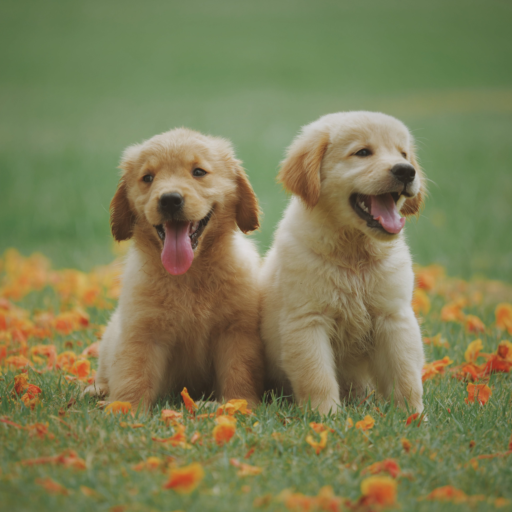} & \imagepng{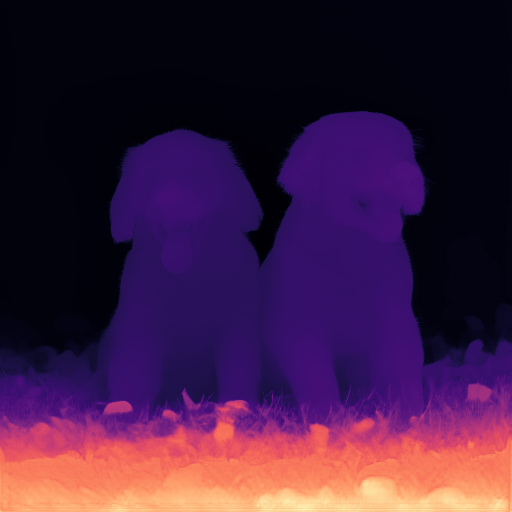} & \imagepng{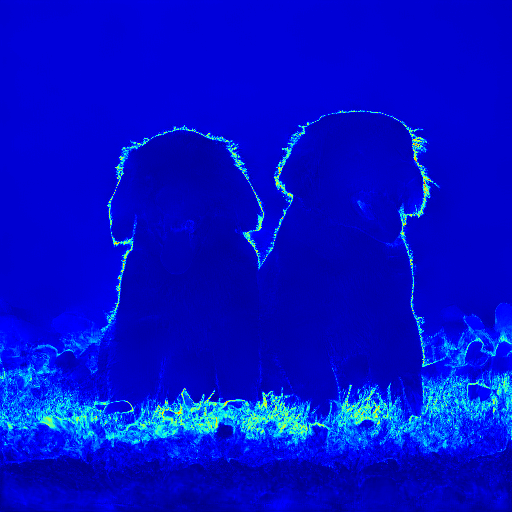} \\
    \end{tabular}
    \caption{Uncertainty in absolute depth produced by the stochasticity in our model.
    \textit{Left}: Original image. \textit{Middle}: Mean depth prediction over ensemble members. \textit{Right}: Standard deviation over ensemble members.}
    \label{fig:uncertainty}
\end{figure}

Another unique feature of our DepthFM model is its inherent ability to provide ensembles of depth predictions, due to the stochastic nature within the generative training paradigm. In addition to improving overall performance, ensembling also provides a robust method for quantifying confidence or uncalibrated uncertainty. We estimate the confidence of a prediction by calculating the standard deviation of predictions across ensemble members. A higher standard deviation implies that the model's predictions are less consistent and more sensitive to the stochasticity present in our model.
\Cref{fig:uncertainty} shows an example image, the corresponding depth estimate, and the uncalibrated uncertainty. The ensemble members show noticeable differences, especially in the high frequency regions. Given the drastic depth contrasts within these regions, the variance along the edges again highlights the ability of our model to sample from a reasonable posterior distribution.
We evaluate the correlation between uncertainty and depth prediction accuracy on the Middlebury-2014 dataset~\cite{scharstein2014high}. Using a threshold to identify regions of high and low uncertainty, we find that the L1 loss is, on average, $3.21$ times higher in high-uncertainty regions, indicating a strong correlation between uncertainty and error. Notably, uncertainty does not impact the fidelity of individual ensemble members. Regardless of variations, ensemble members and their mean consistently produce high-fidelity predictions, as shown in \Cref{fig:fidelity-zoom-ins} and \Cref{tab:precision-recall}. Additional visualizations are provided in the appendix.

\begin{table}
    \centering
    \footnotesize
    \begin{tabular}{lcc}
    \toprule
    Method & EP (\%) $\uparrow$ & ER (\%) $\uparrow$  \\
        \midrule
        Depth Anything \tiny~\cite{yang2024depthanything}
            & 29.32 & 29.80 \\
        Depth Anything v2  \tiny~\cite{yang2024depthanythingv2}
            & 31.67 & 40.25 \\
        \grayrule
        DepthFM \tiny(\textit{Ours})
            & \textbf{33.54} & \textbf{49.31} \\
    \bottomrule
    \end{tabular}
     \caption{We measure zero-shot edge precision (EP) and recall (ER) on the high-resolution RGBD Middlebury-2014 dataset. Our method excels at high detail.
     }
    \label{tab:precision-recall}
\end{table}

\begin{table}
\centering
\footnotesize
\begin{tabular}{lc}
    \toprule
        Method & RMSE $\downarrow$\\
    \midrule
        NLSPN \tiny\cite{NLSPN} &  0.092  \\
        DSN \tiny\cite{DSN} & 0.102  \\ 
        Struct-MDC \tiny\cite{StructMDC} & 0.245  \\ 
        CompletionFormer \tiny\cite{zhang2023completionformer} & 0.090\\
    \grayrule
    DepthFM \tiny\textit{(Ours)} & \textbf{0.077} \\
    \bottomrule
\end{tabular}
\caption{Comparison of \textit{Depth Completion} on NYUv2.}
\label{table:depthcompl}
\end{table}

\begin{table}
    \centering
    \footnotesize
    \begin{tabular}{lcc}
        \toprule
        NFEs & 1 & 10 \\
        \midrule
            noise $\rightarrow$ depth & 92.4 & 92.6 \\
            image $\rightarrow$ depth \tiny(\textit{Ours}) & \textbf{94.6} & \textbf{95.5} \\
        \bottomrule
    \end{tabular}
    \caption{$\delta1$ accuracy for different starting distributions on NYUv2. Direct transport is better than starting from noise.
    }
    \label{tab:starting_dist}
\end{table}

\begin{table}
    \centering
    \footnotesize
    \begin{tabular}{lccc}
    \toprule
         $\delta 1\uparrow$                 & {Scratch} & \text{LoRA} & \text{Finetune} \\ \midrule
    NYUv2   & 80.0    & 75.6       & \textbf{95.5}  \\
    DIODE & 68.8    & 70.0       & \textbf{78.5}  \\ \bottomrule
    \end{tabular}
    \caption{
        Image pre-training and high model capacity are required for high accuracy.
    }
    \label{tab:prior-ablation}
\end{table}

\begin{table}[t]
    \centering
    \footnotesize
    \begin{tabular}{ccc}
    \toprule
     {Image Prior} & {Depth Prior} & $\delta1 \uparrow$\\
    \midrule 
     \no & \no & 80.0\\
     \yes & \no & 95.5 \\
     \yes & \yes & \textbf{96.3} \\  
     \bottomrule
    \end{tabular}
    \caption{
        Including Image and Depth Prior improves zero-shot performance on NYUv2.
    }
    \label{table:external_knowledge}
\end{table}

\subsection{Depth Completion}
\label{sec:depthcompletion}
An important task related to depth estimation is depth completion. Due to hardware limitations of depth sensors, only a partial depth map is usually available. Therefore, the task is to fill in the rest of the missing depth values with suitable depth estimates. Following previous conventions~\cite{NLSPN, DSN}, we fine-tune our DepthFM to complete depth maps where only 2\% of the ground truth pixels are available and evaluate it using the root mean square error (RMSE). \Cref{table:depthcompl} shows that with minimal fine-tuning, DepthFM can achieve state-of-the-art depth completion results on the NYUv2 dataset \cite{silberman12nyuv2}. We provide training details and additional depth completion results in the appendix.

\subsection{Ablation Studies}

\subsubsection{Image-depth Coupling}
We compare DepthFM to a na\"ive Flow Matching (FM) baseline. While na\"ive FM also uses an optimal transport-based objective to regress vector fields, it starts from Gaussian noise with $p(x_0) \sim \mathcal{N}(0, \mathbb{I})$. In contrast, our method starts directly from the latent code of the input image. Both models have access to the image as conditioning information over the entire ODE trajectory. The results in \Cref{tab:starting_dist} demonstrate that starting directly from the latent image representation yields significantly higher accuracy, especially in the low NFE regime.

\begin{figure}[htb]
    \centering
    \includegraphics[width=0.8\linewidth]{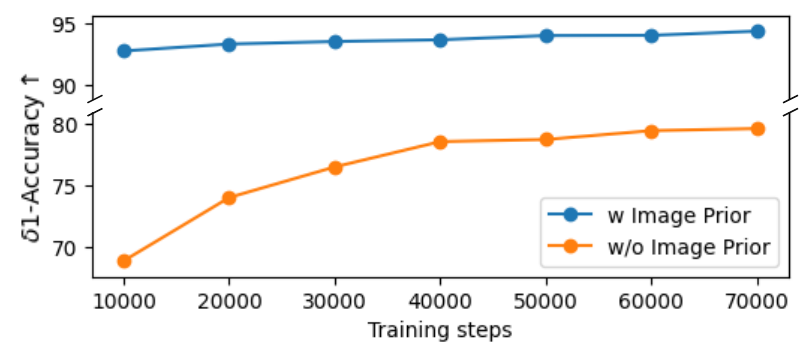}
    \caption{{Image prior can boost the training efficiency and performance on NYUv2.
    } 
     \label{fig:training_efficiency}
    }
\end{figure}

\paragraph{Knowledge Transfer from Image Prior}
We investigate the importance of the image prior and evaluate its impact on performance. We provide metrics for training the same architecture with identical training settings from scratch, from an image prior, and using adapters (LoRA~\cite{hu2021lora}) in~\Cref{tab:prior-ablation}. For LoRA, we use rank 8 and keep the rest of the training details the same.
Our observations indicate that adapters significantly limit the fine-tuning process, as they do not provide enough modeling capacity to transfer the image prior to depth estimation. Training from scratch without fine-tuning does not achieve nearly as good a performance, despite our best efforts to optimize it. Therefore, we conclude that a strong image prior and sufficient modeling capacity is essential to provide important visual cues for depth inference.
In \Cref{fig:training_efficiency}, we further plot the $\delta1$-accuracy versus training steps on NYUv2. Compared with the default baseline, we demonstrate that we can also achieve better training efficiency and performance with the extra image prior.

\paragraph{Knowledge Transfer from Depth Prior}
\label{sec:abl:m3-dose}
Building upon the Image Prior, we further explore the impact of the depth prior. Combining these two knowledge sources leads to significant performance improvements, as demonstrated in \Cref{table:external_knowledge}.
For the optimal mixing coefficient $k$, as referred to in \Cref{eq:mix_dataset}, we ablate it in the appendix. We identify an optimal mix where $k=10\%$ discriminative samples combined with synthetic samples yields the best tradeoff. We find that training on more discriminative samples results in blurrier results indicating a trade-off between accuracy and fidelity.

\paragraph{Data and Training Efficiency}
Our method can achieve a better trade-off between training data and performance by using the external depth and the image prior. Note that our approach differs significantly from Depth Anything~\citep{yang2024depthanything} in two important ways. First, while Depth Anything uses 62M discriminative samples, we achieve optimal results with only 7.4K discriminative samples. Second, Depth Anything requires strong augmentation schemes to take advantage of the pseudo-labeled dataset, while our approach does not require such augmentations.

\paragraph{Noise Augmentation}
Following the notation from variance-preserving diffusion models, we apply noise to the image samples according to the cosine schedule proposed by ~\cite{nichol2021improved}. Through empirical analysis in \Cref{tab:noise_aug}, we determine that a noise augmentation level of $t_s=0.4$ is optimal.

\begin{table}
\centering
\footnotesize
\begin{tabular}{lccccc}
    \toprule
    & \text{0.1}  & \text{0.2}  & \text{0.4}           & \text{0.6}           & \text{0.8}  \\ \midrule
    NYUv2 $\delta 1(\uparrow)$  & 93.7 & 94.4 & \textbf{95.5} & 95.5          & 95.4
    \\ \bottomrule
\end{tabular}
\caption{Noise augmentations level $t_s$.}
\label{tab:noise_aug}
\end{table}

\section{Conclusion}
\label{sec:conclusion}

We present DepthFM, a flow matching approach to generative monocular depth estimation that improves sampling, data fidelity, training, and data efficiency.
First, we improve sampling efficiency by learning a direct transport between image and depth, rather than denoising the Gaussian distribution in depth maps, making our approach faster than current diffusion-based solutions. 
Second, DepthFM provides high-fidelity depth maps without the common artifacts of discriminative depth estimation methods.
Third, we improve training efficiency by using a pre-trained image diffusion model as a prior, providing valuable visual cues to aid depth inference. In addition, we improve data efficiency by using a combination of synthetic data and effectively integrating a discriminative depth prior.

\section*{Acknowledgement}
This project has been supported by the German Federal Ministry for Economic Affairs and Climate Action within the project ``NXT GEN AI METHODS – Generative Methoden für Perzeption, Prädiktion und Planung'', the German Research Foundation (DFG) project 421703927, Bayer AG, and the bidt project KLIMA-MEMES. The authors gratefully acknowledge the Gauss Center for Supercomputing for providing compute through the NIC on JUWELS at JSC and the HPC resources supplied by the Erlangen National High Performance Computing Center (NHR@FAU funded by DFG).

\bibliography{aaai25}

\renewcommand\thesection{\Alph{section}}
\setcounter{section}{0}
\onecolumn

\section{Appendix for DepthFM: Fast Generative Monocular Depth Estimation with Flow Matching}

\begin{figure*}[ht]
    \centering
    \includegraphics[width=\textwidth]{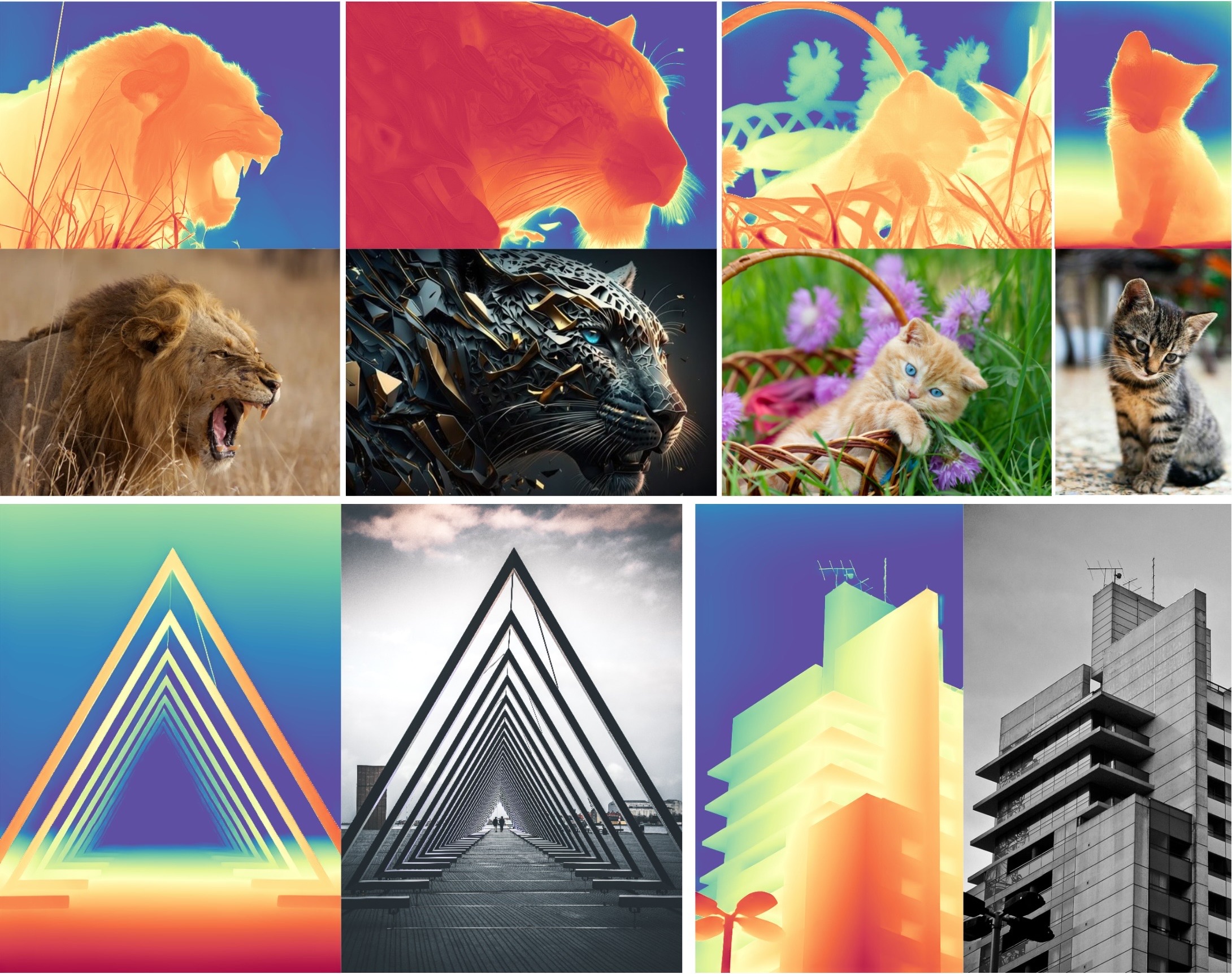}
    \caption{\textbf{DepthFM} generalizes to real-world data with varying resolutions and aspect ratios. Best viewed when zoomed in.}
    \label{fig:multi-resolution}
\end{figure*}

\subsection{Additional Qualitative Comparison}
 
\cref{fig:sup:indoor-comparison} and \cref{fig:sup:outdoor-comparison} show additional qualitative comparisons with other state-of-the-art monocular depth estimation methods. Images and predictions for the other models are taken from \cite{saxena2023dmd}. For both, indoor and outdoor scenes, our model consistently produces depth maps with higher fidelity than the other methods. Our method particularly excels in outdoor scenes, showing a clear distinction between objects in the far distance and the sky, especially visible in the samples from DIODE, KITTI, and virtual KITTI 2 in \cref{fig:sup:outdoor-comparison}. ZoeDepth \cite{bhat2023zoedepth} and DMD \cite{saxena2023dmd} fail to distinguish objects that are further away.

Indoor scenes in \cref{fig:sup:indoor-comparison} are from DIODE \cite{diode_dataset}, Hypersim \cite{roberts2021hypersim}, ibims-1 \cite{koch2020ibims-1}, NYU-v2 \cite{silberman12nyuv2}, and SunRGBD \cite{Song2015_sunrgbd}. Outdoor scenes in~\cref{fig:sup:outdoor-comparison} are from DDAD \cite{packnetDDAD}, DIML \cite{cho2021dimlcvl}, DIODE \cite{diode_dataset}, KITTI \cite{kitti}, and virtual KITTI 2 \cite{cabon2020vkitti2}.

\subsubsection{Generalization across Resolutions} \Cref{fig:multi-resolution} shows a collection of images with varying resolutions and aspect ratios. Even though our model was only trained on synthetic data on a fixed resolution of $384 \times 512$ pixels, it generalizes well to real data with different resolutions and aspect ratios while producing realistic depth maps. 

\subsubsection{Depth Inpainting} 
\begin{figure}
    \centering
    \includegraphics[width=0.6\textwidth]{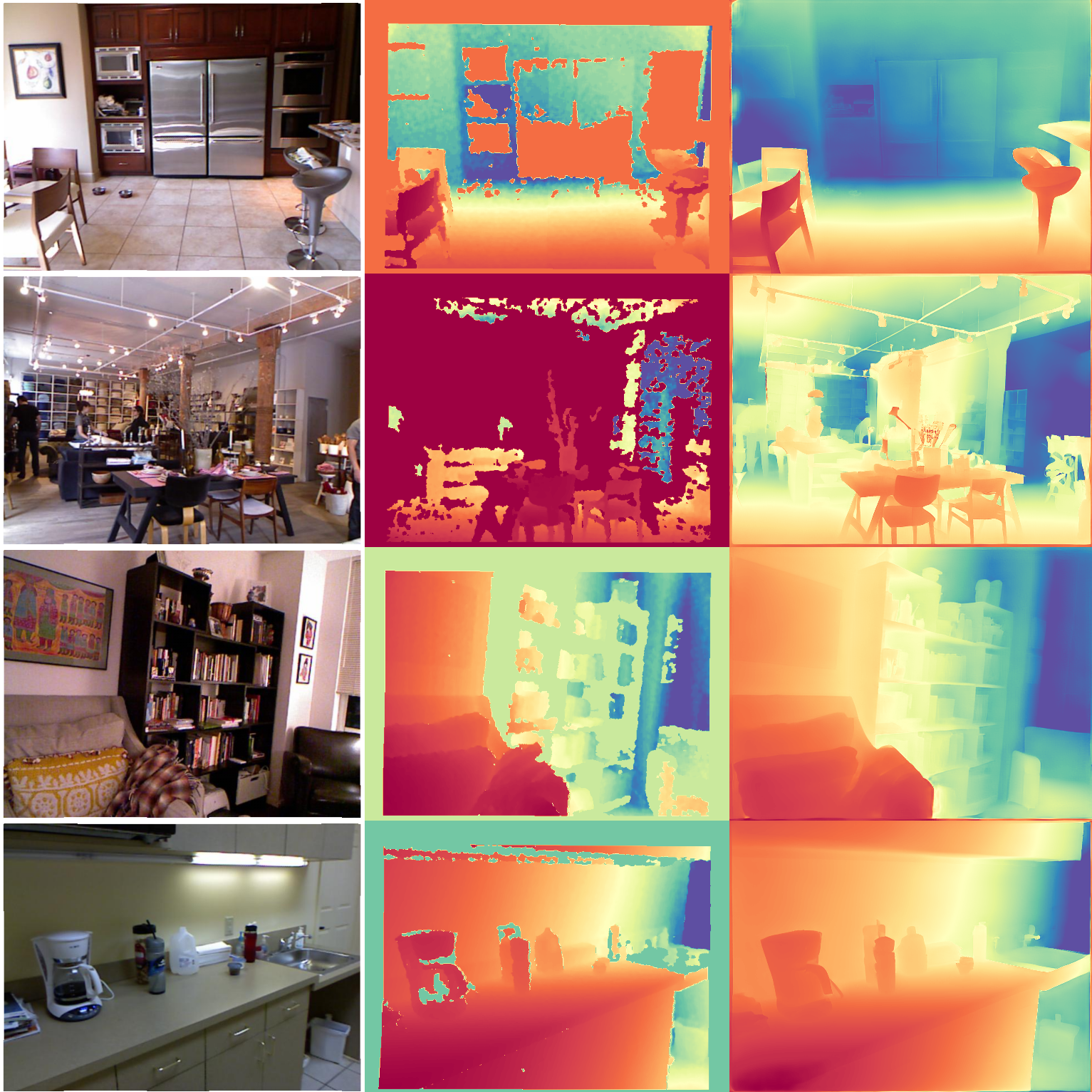}
    \vspace{-2mm}
    \caption{Zero-shot depth completion on the NYUv2 dataset \cite{silberman12nyuv2}. The middle column is the ground truth raw depth, and the rightmost column is our depth prediction. }
    \label{fig:zeroshot-depth-compl}
\end{figure}

\Cref{fig:depthcompl} presents qualitative results for the depth inpainting task. Our method successfully inpaints depth from a highly sparse partial depth input. We show qualitatively results in~\Cref{fig:zeroshot-depth-compl} when we apply our model on unseen dataset NYUv2.

\subsubsection{Edge Fidelity}
\begin{figure*}[t]
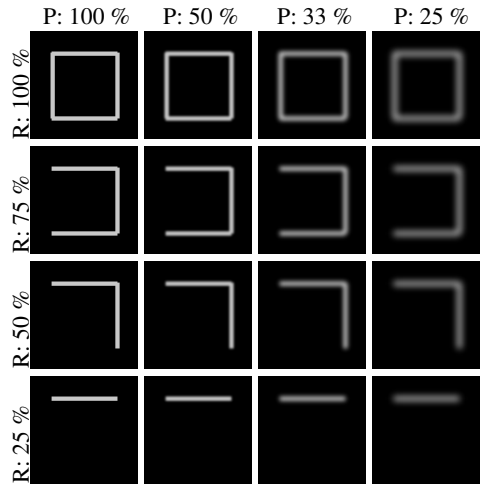

    \setlength\tabcolsep{0.2pt}
    \center
    \small
    \newcommand{\imgwidth}{0.08\textwidth}    
    \newcommand{\imagepng}[2]{
    \includegraphics[width=\imgwidth]{figs/supp/edges_toy/r#1_l#2.png}
    }
    \newcommand{\rowedgestoy}[1]{
        \imagepng{0}{#1} & \imagepng{1}{#1} & \imagepng{2}{#1} & \imagepng{3}{#1}
    }

    \begin{tabular}{ccccc}
        & P: 100 \% & P: 50 \% & P: 33 \% & P: 25 \%  \\
        \rotatebox{90}{R: 100 \%}
            & \rowedgestoy{4} \\
        \rotatebox{90}{R: 75 \%}
            & \rowedgestoy{3} \\
        \rotatebox{90}{R: 50 \%}
            & \rowedgestoy{2} \\
        \rotatebox{90}{R: 25 \%}
            & \rowedgestoy{1} \\            
    \end{tabular}
    \caption{Illustration of edge precision and recall. From left to right, blurrier edges result in less precision. From top to bottom, fewer edges result in lower recall. Values reflect real precision and recall for the respective case.}
    \label{fig:sup:edges_toy}
\end{figure*}

\begin{figure*}[t]
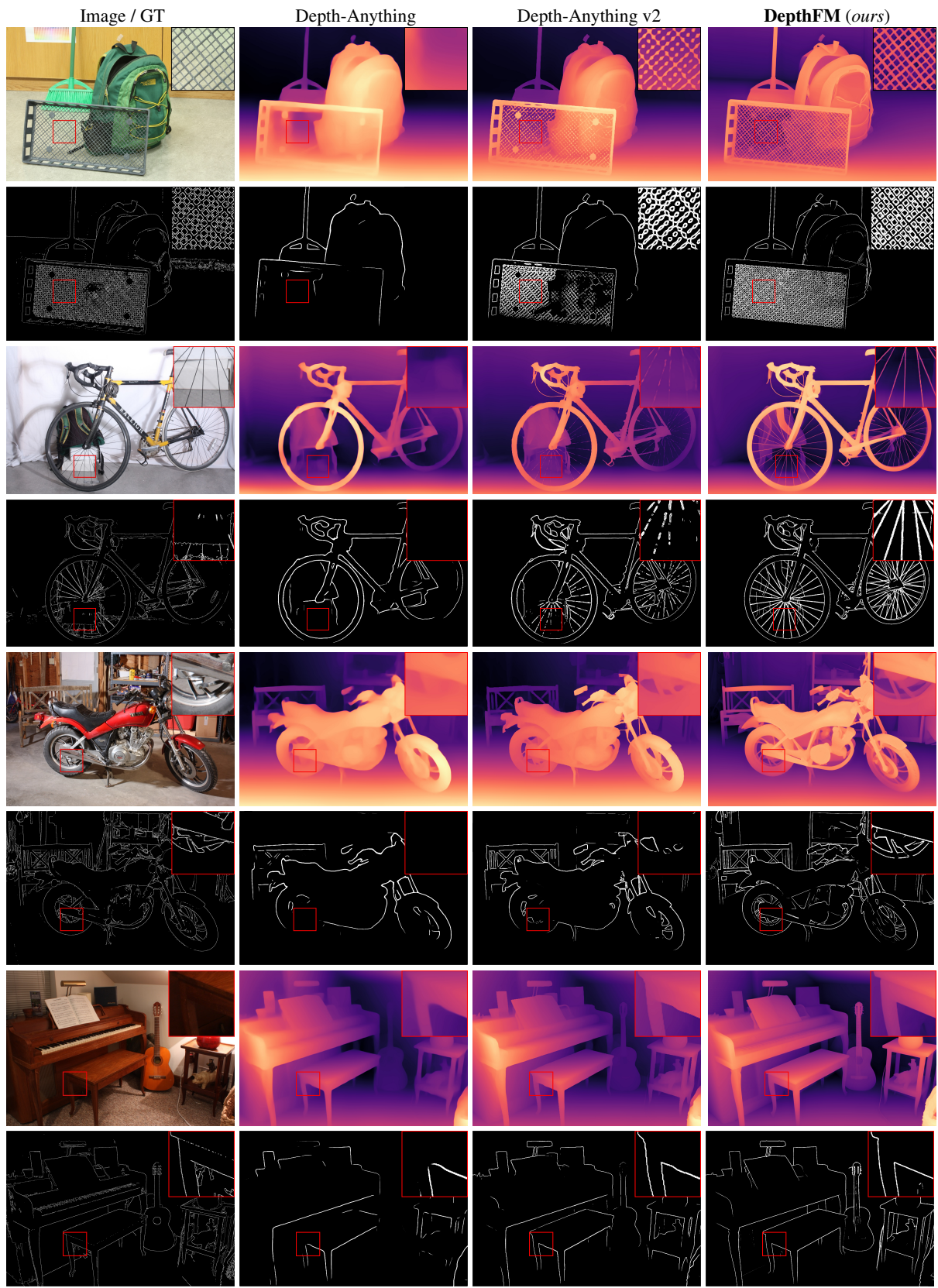

    \setlength\tabcolsep{0.2pt}
    \center
    \small
    \newcommand{\imgwidth}{0.22\textwidth}
    
    \newcommand{\imagepng}[2]{
    \includegraphics[width=\imgwidth]{figs/supp/edges_comparison/#1/#2}
    }
        
    \newcommand{\rowdepth}[1]{
        \imagepng{#1}{image.jpg} &
        \imagepng{#1}{dav1.png} &
        \imagepng{#1}{dav2.png} &
        \imagepng{#1}{ours.png}
    }
    
    \newcommand{\rowedge}[1]{
        \imagepng{#1}{gt_edge.png} &
        \imagepng{#1}{dav1_edge.png} &
        \imagepng{#1}{dav2_edge.png} &
        \imagepng{#1}{ours_edge.png}
    }

    \newcommand{\rows}[1]{
        \rowdepth{#1} \\
        \rowedge{#1}
    }
    
    \begin{tabular}{cccc}
        Image / GT & Depth-Anything & Depth-Anything v2 & \textbf{DepthFM} (\textit{ours}) \\
        \rows{Backpack-imperfect-0} \\
        \rows{Bicycle1-imperfect-1} \\
        \rows{Motorcycle-imperfect-0} \\
        \rows{Piano-imperfect-0} \\
    \end{tabular}
    \caption{Qualitative results comparing DepthFM to counterpart discriminative models on the high-resolution Middlebury-2014 dataset, along with extracted edges. DepthFM consistently produces sharper depth predictions overall, which is also reflected in the quantitative metrics reported in the main paper. Note that some pixels in the Middlebury-2014 ground truth depth maps are invalid, which is why certain details (e.g., the spokes in the bicycle image) are missing. However, our model is able to accurately extract these details with high fidelity. Best viewed when zoomed in.}
    \label{fig:sup:edges_comparison}
\end{figure*}

In the main paper, we quantitatively demonstrate the higher fidelity of DepthFM compared to other discriminative depth estimation methods, following the approach of \cite{Hu2018RevisitingSI}. In particular, we extract edges from the ground truth and predicted depth and measure edge precision and recall on the Middlebury-2014 dataset \cite{scharstein2014high}. Sharp and precise edges result in a high precision, whereas blurry edges yield a low precision. Recall reflects the overall accuracy of correctly detecting an edge. We schematically illustrate this in \Cref{fig:sup:edges_toy}. \Cref{fig:sup:edges_comparison} directly compares our method with the state-of-the-art discriminative depth estimation model family from Depth-Anything. The Depth-Anything models suffer from missing details, especially visible in the spokes of the bicycle or the pattern of the box in front of the backpack. In contrast, DepthFM is able to show these fine-grained patterns, highlighting its versatility. Also, while Depth-Anything v2 shows curvature along the edge of the piano chair, DepthFM correctly shows straight edges.  

\subsubsection{High Fidelity in Ensemble Mean and Individual Ensemble Members}
\begin{figure*}[t]
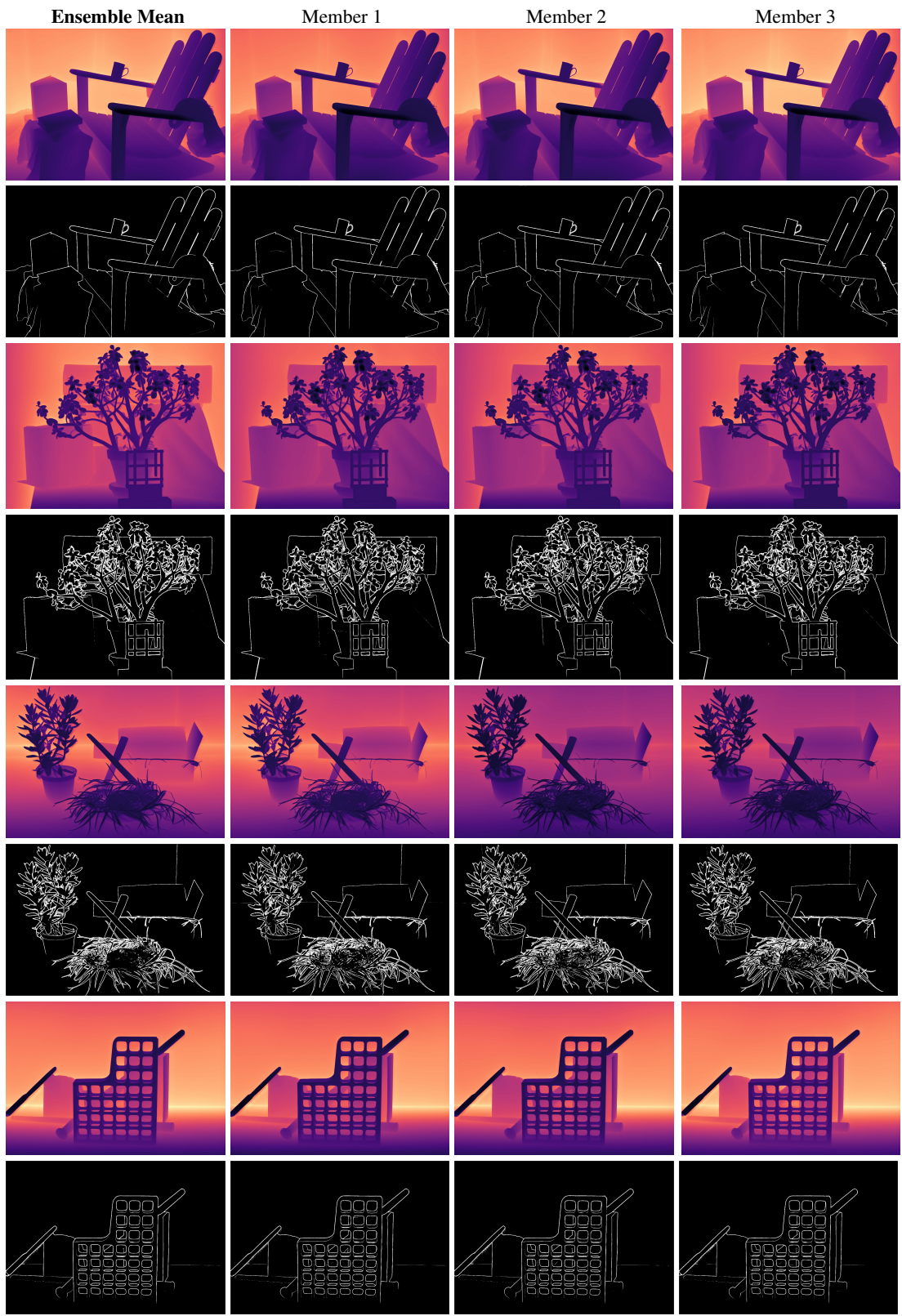

    \setlength\tabcolsep{0.2pt}
    \center
    \small
    \newcommand{\imgwidth}{0.2\textwidth}
    
    \newcommand{\imagepng}[2]{
    \includegraphics[width=\imgwidth]{figs/supp/ensemble_edges/#1_#2.jpg}
    }
        
    \newcommand{\rowy}[2]{
        \imagepng{#1}{#2_ensemble} &
        \imagepng{#1}{#2_0} &
        \imagepng{#1}{#2_1} &
        \imagepng{#1}{#2_2}
    }

    \newcommand{\fullrow}[1]{
        \rowy{#1}{depth} \\
        \rowy{#1}{edge}
    }
    
    \begin{tabular}{cccc}
        \textbf{Ensemble Mean} & Member 1 & Member 2 & Member 3 \\
        \fullrow{Adirondack-perfect} \\
        \fullrow{Jadeplant-perfect} \\
        \fullrow{Sword1-perfect} \\
        \fullrow{Sword2-perfect} \\
    \end{tabular}
    \caption{DepthFM predictions on the high-resolution Middlebury-2014 dataset, along with extracted edges. The edges from both the ensemble mean and the individual ensemble members demonstrate consistency, reflecting the model's ability to maintain high-quality edge details across different predictions. We operate with an ensemble size of 8 but visualize only 3 random members due to space limitations. Best viewed when zoomed in.}
    \label{fig:ensemble-fidelity-edges}
\end{figure*}

\Cref{fig:ensemble-fidelity-edges} qualitatively demonstrates that both the ensemble mean and each individual ensemble member exhibit high-fidelity depth estimations. This is further supported by the consistency of the edges extracted from the mean and the individual ensemble members, all of which maintain sharp and high-fidelity edges.

\begin{figure*}[t]
    \setlength\tabcolsep{0.01pt}
    \center
    \small
    \newcommand{\imgwidth}{0.17\textwidth}
    
    \newcommand{\imagepng}[3]{
    \includegraphics[width=\imgwidth]{figs/supp/ims-depth-comparison/#1/#2_#3.png}
    }
    
    \newcommand{\imagejpg}[3]{
    \includegraphics[width=\imgwidth]{figs/supp/ims-depth-comparison/#1/#2_#3.jpg}
    }
        
    \newcommand{\rowrgbjpg}[2]{
    \imagejpg{#1}{#2}{image} & \imagepng{#1}{#2}{gt} & \imagepng{#1}{#2}{depth_zd} & \imagepng{#1}{#2}{depth_ours} & \imagepng{#1}{#2}{depthfm}
    }
    \begin{tabular}{cccccc}
        & Image & GT & ZoeDepth & DMD & \textbf{Ours} \\
        \rotatebox{90}{\hspace{0.7em} DIODE} &
        \rowrgbjpg{diode_indoor}{00187} \\
        \rotatebox{90}{\hspace{0.4em} Hypersim} &
        \rowrgbjpg{hypersim}{03499} \\
        \rotatebox{90}{\hspace{0.7em} ibims-1} & 
        \rowrgbjpg{ibims}{00027} \\
        \rotatebox{90}{\hspace{1.1em} NYU} & 
        \rowrgbjpg{nyu}{00234} \\
        \rotatebox{90}{\hspace{0.05em} SunRGBD} & 
        \rowrgbjpg{sunrgbd}{01352} \\
    \end{tabular}
    \caption{Indoor scenes comparison. Best viewed when zoomed in.}
    \label{fig:sup:indoor-comparison}
\end{figure*}

\begin{figure*}[t]
    \setlength\tabcolsep{0.2pt}
    \center
    \small
    \newcommand{\imgwidth}{0.17\textwidth}
    
    \newcommand{\imagepng}[3]{
    \includegraphics[width=\imgwidth]{figs/supp/ims-depth-comparison/#1/#2_#3.png}
    }
    
    \newcommand{\imagejpg}[3]{
    \includegraphics[width=\imgwidth]{figs/supp/ims-depth-comparison/#1/#2_#3.jpg}
    }
        
    \newcommand{\rowrgbjpg}[2]{
    \imagejpg{#1}{#2}{image} & \imagepng{#1}{#2}{gt} & \imagepng{#1}{#2}{depth_zd} & \imagepng{#1}{#2}{depth_ours} & \imagepng{#1}{#2}{depthfm}
    }
    
    \begin{tabular}{cccccc}
        & Image & GT & ZoeDepth & DMD & \textbf{Ours} \\
        \rotatebox{90}{\hspace{0.6em} DDAD} &
        \rowrgbjpg{ddad}{02920} \\
        \rotatebox{90}{\hspace{0.4em} DIML} &
        \rowrgbjpg{diml_outdoor}{00292} \\
        \rotatebox{90}{\hspace{0.7em} DIODE} & 
        \rowrgbjpg{diode_outdoor}{00346} \\
        \rotatebox{90}{KITTI} & 
        \rowrgbjpg{kitti}{00166} \\
        \rotatebox{90}{\hspace{-0.4em} vKITTI} & 
        \rowrgbjpg{vkitti2}{00371} \\
    \end{tabular}
    \vspace*{-0.2cm}
    \caption{Outdoor scenes comparison. Best viewed when zoomed in.}
    \label{fig:sup:outdoor-comparison}
\end{figure*}

\subsection{Further Applications}

\subsubsection{Video Depth Estimation} 
\begin{figure*}[t]
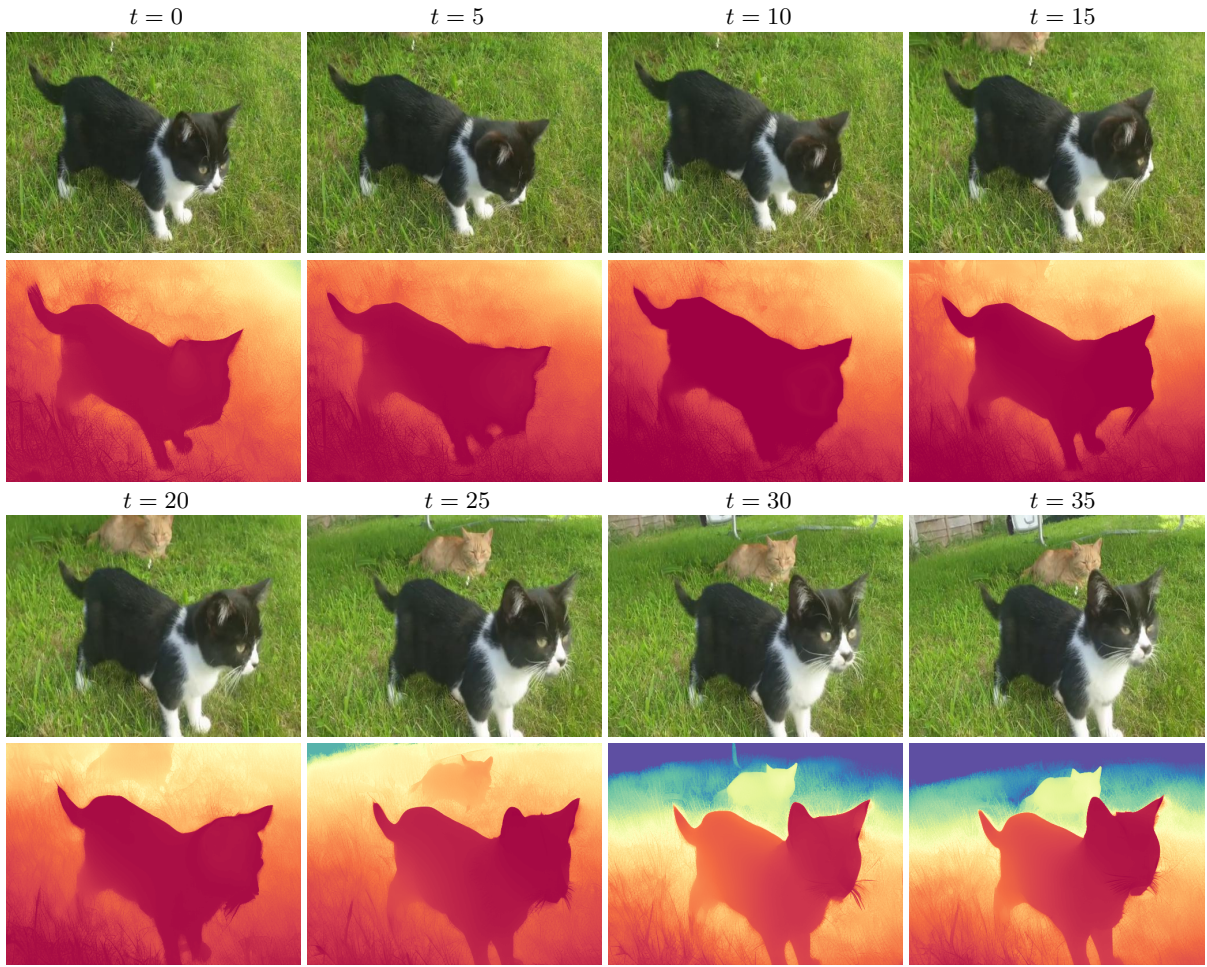

    \setlength\tabcolsep{0.2pt}
    \center
    \small
    \newcommand{\imgwidth}{0.22\textwidth}
    
    \newcommand{\imagepng}[2]{
    \includegraphics[width=\imgwidth]{figs/supp/video/#1_#2}
    }
        
    \newcommand{\rowvideo}[5]{
        \imagepng{#1}{#2} &
        \imagepng{#1}{#3} &
        \imagepng{#1}{#4} &
        \imagepng{#1}{#5}
    }
    
    \begin{tabular}{cccc}
        $t=0$ & $t=5$ & $t=10$ & $t=15$ \\
        \rowvideo{image}{15.jpg}{20.jpg}{25.jpg}{30.jpg} \\
        \rowvideo{depth}{15.png}{20.png}{25.png}{30.png} \\
        $t=20$ & $t=25$ & $t=30$ & $t=35$ \\
        \rowvideo{image}{35.jpg}{40.jpg}{45.jpg}{50.jpg} \\
        \rowvideo{depth}{35.png}{40.png}{45.png}{50.png} \\
    \end{tabular}
    \vspace*{-0.2cm}
    \caption{Video depth prediction with our DepthFM model.}
    \label{fig:sup:video}
\end{figure*}

\cref{fig:sup:video} qualitatively shows our DepthFM model applied on a video. We estimate the depth with $5$ ensemble members and two ODE steps. Since our model predicts relative depth we scale and shift the current frame based on the depth of the previous frame to ensure temporal consistency.

\begin{figure*}[t]
    \centering
    \includegraphics[width=\textwidth]{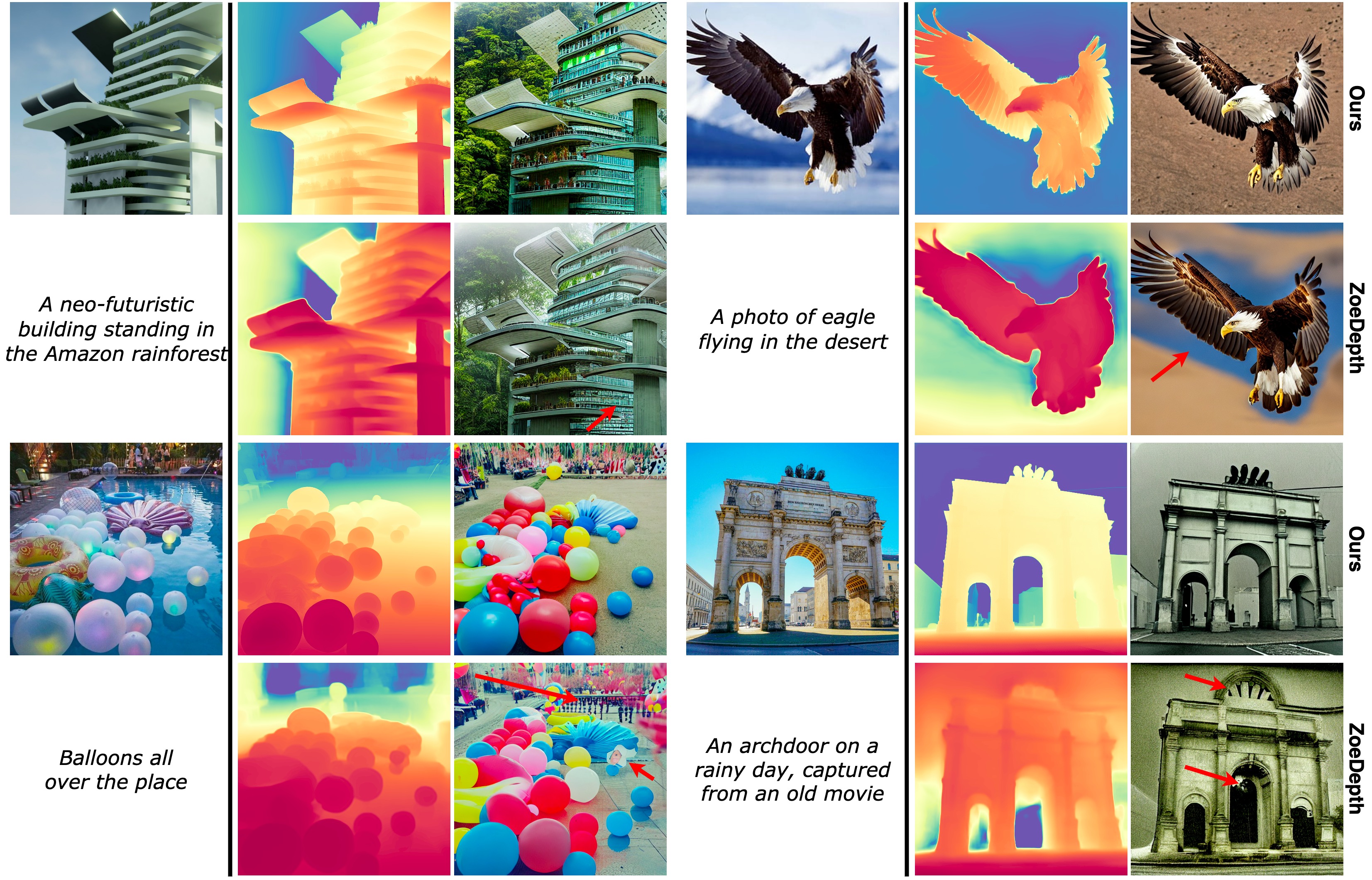}
    \caption{The superior and high-fidelity depth map generation capability of DepthFM also empowers ControlNet \cite{zhang2023adding} to generate images with depth fields that closely mimic the source image in a zero-shot manner. ZoeDepth \cite{bhat2023zoedepth} fails to replicate the exact depth field, as evidenced by the discrepancy between the corresponding ControlNet output and the source images, highlighted by the red arrows. The prompts for ControlNet are visualized in the lower left corners, and the images are generated using the same random seed. Best viewed when zoomed in.}
    \label{fig:controlnet}
\end{figure*}

\subsubsection{Conditional Synthesis}

\cref{fig:controlnet} shows a comparison of depth-conditioned synthesis results. We first infer depth based on an image using our DepthFM and the ZoeDepth \cite{bhat2023zoedepth} model, and then use a pre-trained depth-to-image ControlNet \cite{zhang2023adding} with Stable Diffusion 1.5 to synthesize new samples based on the depth maps and a text prompt. We can clearly observe that the depth maps obtained with ZoeDepth do not reflect the actual depth well and additionally are inaccurate for some parts of the image. In contrast, our method yields sharp and realistic depth maps. This result is also reflected in the synthesized results, where images created based on our depth map more closely resemble the actual image.

\subsection{Additional Ablations}
\label{sec:sup:ablations}

\paragraph{Pseudo Data Ratio} We find that there exists an optimal pseudo data ratio, as shown in~\Cref{supp:pseudo_ratio}. A ratio with too few depth prior samples fails to maximize performance, while an excess of depth prior samples results in a decline in both fidelity and metrics, as evidenced in~\Cref{fig:m3-dose}.

\begin{table}
    \centering
    \begin{tabular}{lcccc}
    \toprule
    {\textbf{k}} & 0.0  & 0.1  & 1.0 \\ 
    \midrule
    {NYUv2}    & 95.5 & 96.3 & 96.7 \\
    {DIODE}    & 78.5 & 80.0 & 81.0 \\
         \bottomrule
    \end{tabular}
    \caption{Ablation about the number of depth prior samples used in our training pipeline on $\delta1 (\uparrow)$. $k$ represents the ratio of the number of depth prior samples to the number of GT samples. While an increased ratio results in better performance, the images get increasingly blurry.}
    \label{supp:pseudo_ratio}
\end{table}

\begin{figure}
    \centering
    \includegraphics[scale=0.6]{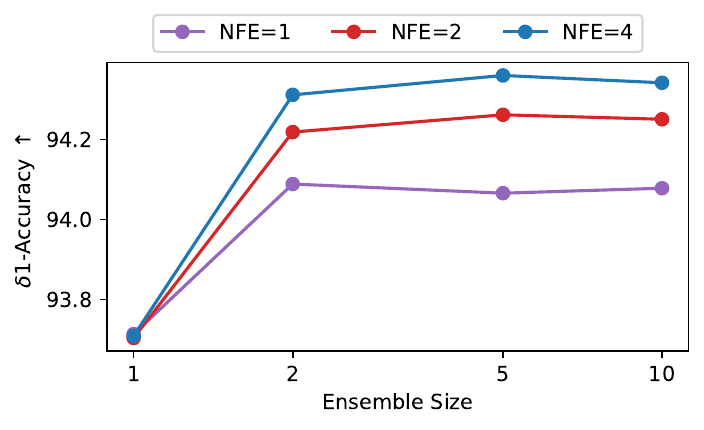}
    \caption{Ablation of ensemble size for different number of function evaluations (NFE) on the NYUv2 dataset.}
    \label{fig:ensembles-nfe}
\end{figure}

\paragraph{Ensemble} Marigold uses ensembling techniques to further improve depth estimation. We find that this also improves performance in our case, as visualized in \Cref{fig:ensembles-nfe}. 

\subsubsection{Data Preprocessing}
\label{sec:sup:datapreproc}
\begin{figure*}
    \centering
    \includegraphics[width=.8\textwidth]{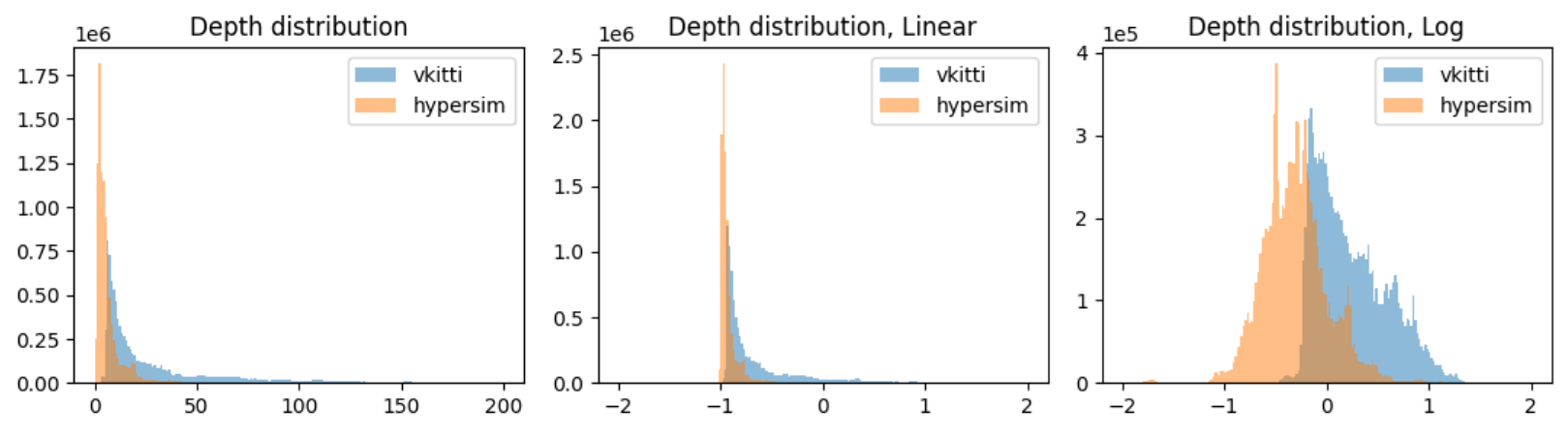}
    \caption{The distribution of the raw metric depth, and the sample distribution after different normalization techniques.}
    \label{fig:normalization}
\end{figure*}

The pre-trained autoencoder of latent diffusion models \cite{rombach2022high} expects the input data to be in the range $[-1, 1]$. To reach this range, we ablate two normalization techniques. First, linear normalization, which linearly normalizes the values at the 2\% and 98\% percentiles. The second is logarithmic normalization, which shifts the percentile normalization to logarithmic space. The raw depth and the corresponding normalized distributions are visualized in \cref{fig:normalization}. In particular, linear normalization allocates limited representational capacity to indoor scenes, whereas log-scaled depth allocation alleviates the problem and allocates similar capacity to both indoor and outdoor scenes. We also observe empirical benefits from log scaling, as shown in \cref{tab:normalization_metrics}.

It is worth noting that some of the depth data, including the synthetic data on which we train our models, contains invalid values, e.g. due to infinite distance or lack of valid scene geometry at certain pixels. We compute the invalid mask from the depth maps and resize it to fit the smaller latent space. We then threshold the small invalid depth map to map all values below 1 to ensure we only get valid ground truth depth maps within the mask. The invalid values are interpolated using nearest neighbors and then passed to the encoder to ensure compatibility with a valid latent representation.

\begin{table}[]
    \centering
    \setlength\tabcolsep{6pt}
    \begin{tabular}{lcccc}
    \toprule
        & \multicolumn{2}{c}{NYU-v2} 
        & \multicolumn{2}{c}{DIODE } \\
    Transform
    & AbsRel$\downarrow$    & $\delta1$ $\uparrow$          & AbsRel$\downarrow$    & $\delta1$ $\uparrow$ \\
    \midrule
    Identity
    & 0.080                 & 93.9                         & 0.237                 & 78.4                 \\
    Log
    & \textbf{0.055}        & \textbf{96.3}                & \textbf{0.212}        & \textbf{80.0}         \\
    \bottomrule
    \end{tabular}
    \caption{Ablation of the data normalization strategy.}
    \label{tab:normalization_metrics}
\end{table}

\subsubsection{Training Data} 
\begin{figure*}[t]
    \centering
    \includegraphics[width=0.7\textwidth]{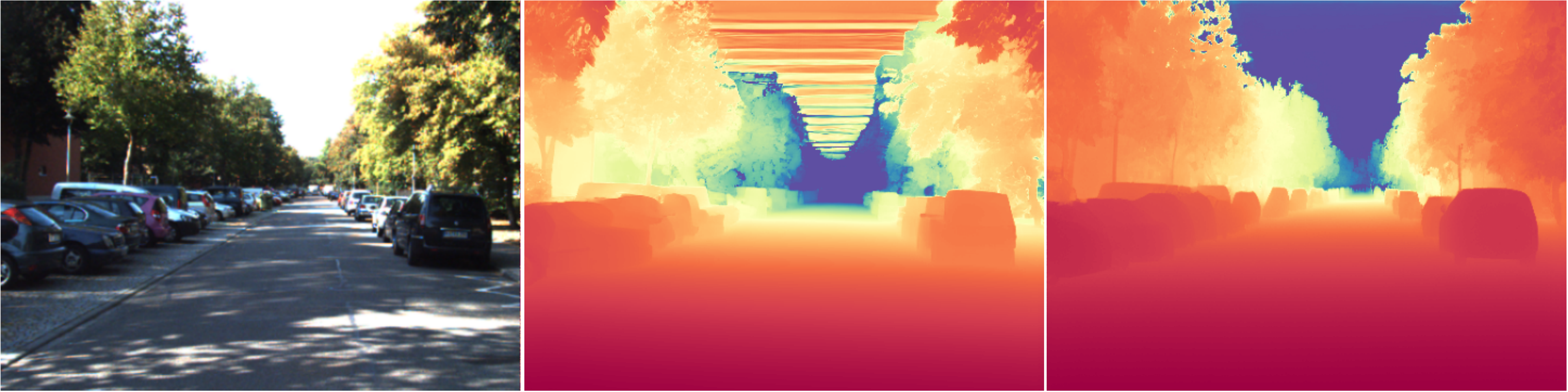}
    \caption{Comparison of zero-shot depth estimation on the KITTI dataset. The left image is the groundtruth image, the middle one showcases the model trained only on Hypersim, and the right image displays the model trained on a combination of Hypersim and VirtualKitti. This comparison underscores the extra value of incorporating synthetic outdoor training data for improved performance in outdoor scenes.}
    \label{fig:vkitti-comparison}
\end{figure*}

Similar to \textit{Marigold} \cite{ke2023marigold}, we train our model on the synthetic Hypersim \cite{roberts2021hypersim} and virtual KITTI \cite{cabon2020vkitti2} datasets. \cref{fig:vkitti-comparison} shows the zero-shot performance comparison on the KITTI dataset between two models: one trained only on the indoor Hypersim dataset and another trained on a combination of both datasets. The results clearly show that the integration of v-KITTI into the training process improves the inference accuracy for outdoor scenes and provides realistic depth estimates even for distant objects.

\begin{figure*}
    \centering
    \setlength\tabcolsep{0.01pt}
    \center
    \small
    \newcommand{\imgwidth}{0.19\textwidth}
    \newcommand{\imagepng}[2]{
        \includegraphics[width=\imgwidth]{figs/supp/ratio_k/#1_#2.jpg}
    }
    \newcommand{\rowpng}[1]{
        \imagepng{#1}{original} & \imagepng{#1}{hv} & \imagepng{#1}{ratio10} & \imagepng{#1}{ratio100}
    }
    \begin{tabular}{cccc}
        Image & Image Prior & +Depth Prior ($k=0.1$)  & +Depth Prior ($k=1.0$) \\
        \rowpng{00} \\
        \rowpng{01} \\
    \end{tabular}
    \caption{We demonstrate qualitatively that an excessive number of discriminative samples will improve metric performance but also reduce the fidelity of depth predictions. $k$ represents the ratio between the depth prior samples and the GT samples. The optimal ratio of 0.1 corresponds to a balanced mix of 74K depth prior samples and 7.4K GT samples.
    }
    \label{fig:m3-dose}
\end{figure*}

\subsection{Training Details}
\label{sec:training_details}
For all of the training runs of both monocular depth estimation and the depth inpainting model, we finetune from the SD2.1 image prior model. Consequently, we keep the network UNet architecture fixed except for the first convolutional layer to incorporate the extra image conditioning. We duplicate channel-wise the weights of the first convolutional layer and tune down the weights of that layer to preserve the input magnitude to the subsequent layer. We deactivate text conditioning and employ null-text embedding as the input for cross-attention.
We employ a global batch size of $128$, a learning rate of $3\times 10^{-5}$, and an EMA rate of $0.999$. In all flow matching paradigms we set $\sigma_\text{min}$ as $1\mathrm{e}{-8}$.
Since our model is trained on synthetic data only, we use the affine-invariant loss to ignore the scale and shift of each unknown sample during evaluation. We perform a least-squares fit of the scale and shift to our model's prediction $d^*$ in the corresponding normalization space to minimize the MSE, such that $\hat{d} = \frac{d^*-t(d^*)}{s(d^*)}$.
We conduct all experiments on NVIDIA A100 40GB GPUs and use PyTorch version 2.1.0. Other Python dependencies are included in the GitHub repository.

\subsubsection{Depth Normalization}
\label{sec:norm} 
Inspired by~\cite{ke2023marigold}, we convert the depth images into three channels to simulate an RGB image. We also find that the data range is essential for the performance of the latent encoder. While images can be normalized quite easily to a range of [-1, 1], we need to normalize the depth images also to the same value range. Similar to \cite{ke2023marigold}, we compute the quantiles of the individual datasets and normalize them individually. The normalization process per depth image $d \in \mathbb{R}^{ H \times W}$ can be depicted as
\begin{equation}
    \Tilde{d} = \left(
    \frac{\texttt{Fn}(d) - \texttt{Fn}(d_{2})}{\texttt{Fn}(d_{98}) - \texttt{Fn}(d_{2})} - 
    0.5\right) \cdot 2,
\end{equation}
where $d_2$ and $d_{98}$ correspond to the 2\% and the 98\% quantile of the dataset, and \texttt{Fn} is the normalization function, where we can use the \texttt{Log} or \texttt{Linear} operation.
After this normalization function, we pass it through the encoder and obtain the corresponding latent representations. As shown in \Cref{tab:normalization_metrics}, we observe an additional performance improvement when using \texttt{Log} normalization.

\begin{figure*}[t]
    \center \small
    \setlength\tabcolsep{0.1pt}
    \newcommand{\imgwidth}{0.2\textwidth}
    
    \begin{tabular}{ccc}
        Partial Depth & Depth Prediction & Ground Truth \\
        {
        \includegraphics[width=\imgwidth]{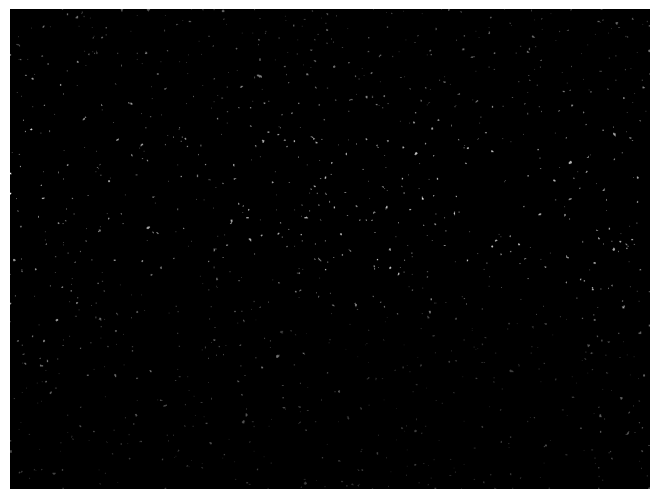}
        } & {
        \includegraphics[width=\imgwidth]{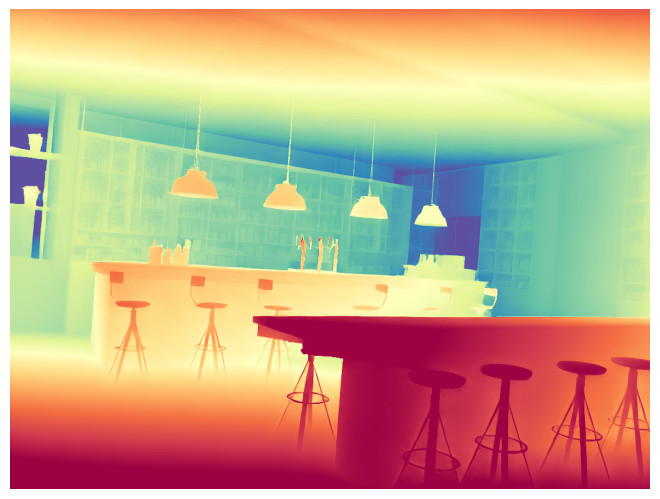}
        } & {
        \includegraphics[width=\imgwidth]{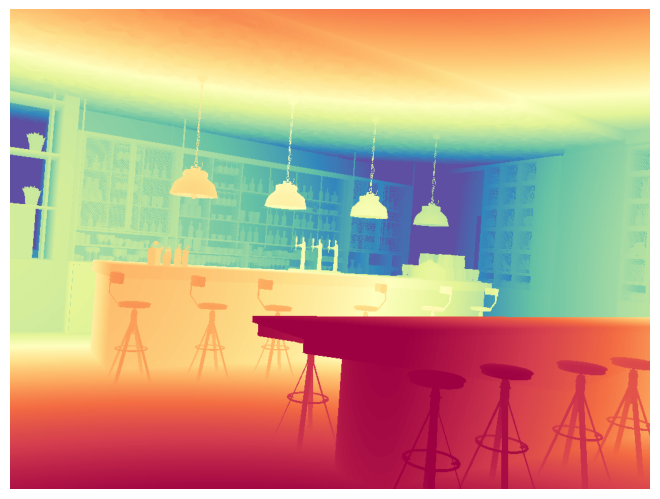}
        }\\
        {
        \includegraphics[width=\imgwidth]{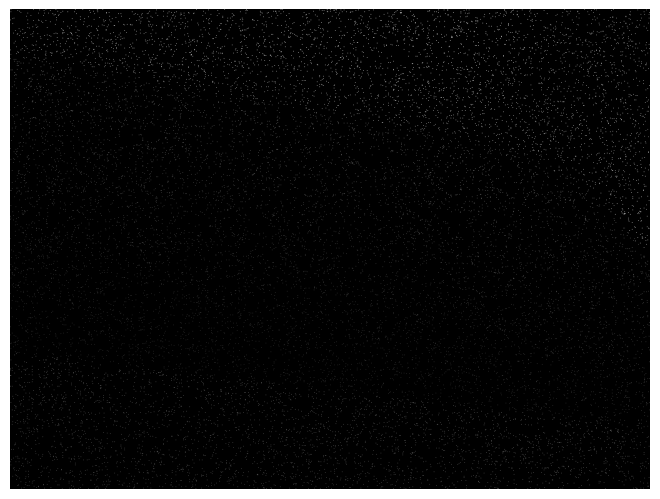}
        } & {
        \includegraphics[width=\imgwidth]{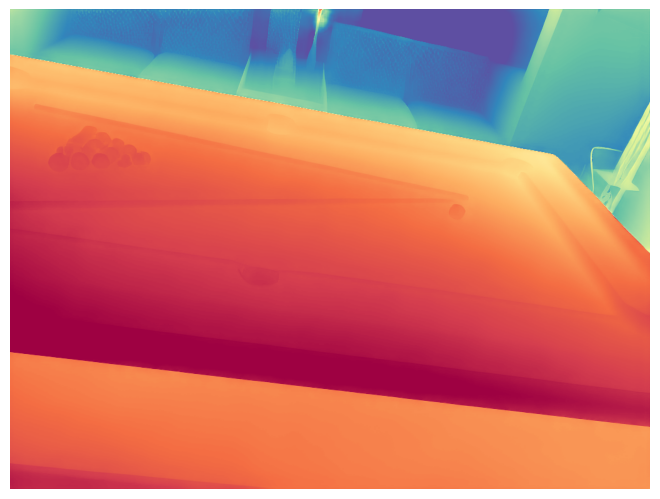}
        } & {
        \includegraphics[width=\imgwidth]{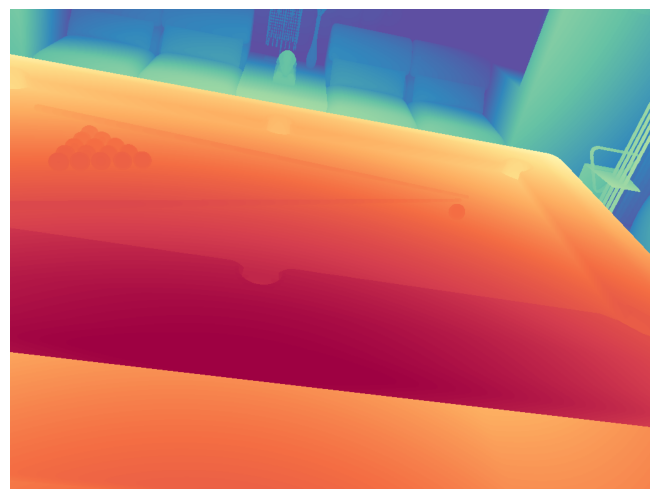}
        }\\
        {
        \includegraphics[width=\imgwidth]{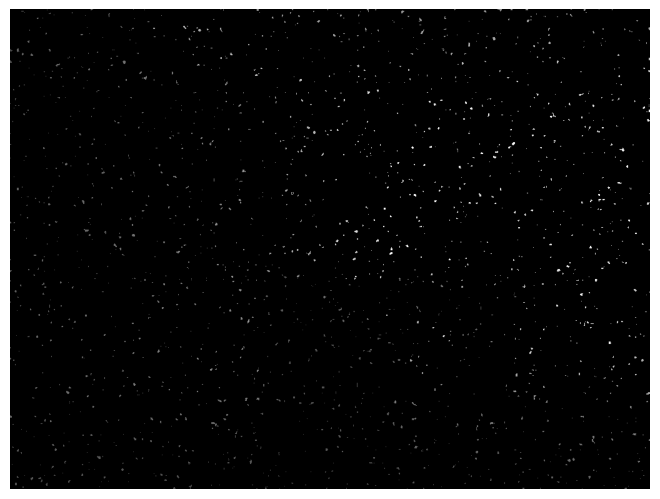}
        } & {
        \includegraphics[width=\imgwidth]{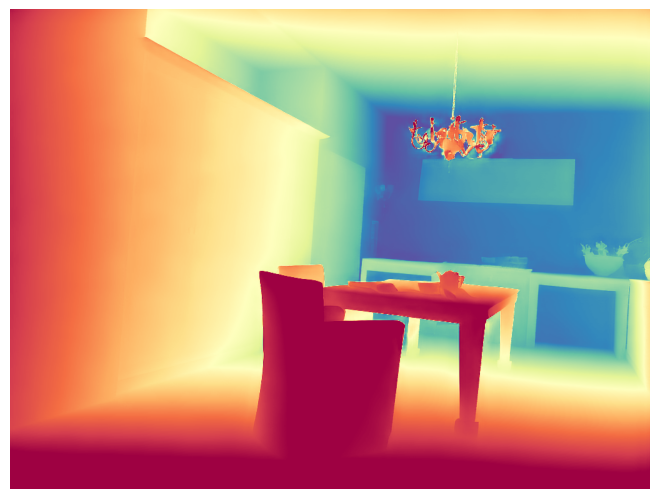}
        } & {
        \includegraphics[width=\imgwidth]{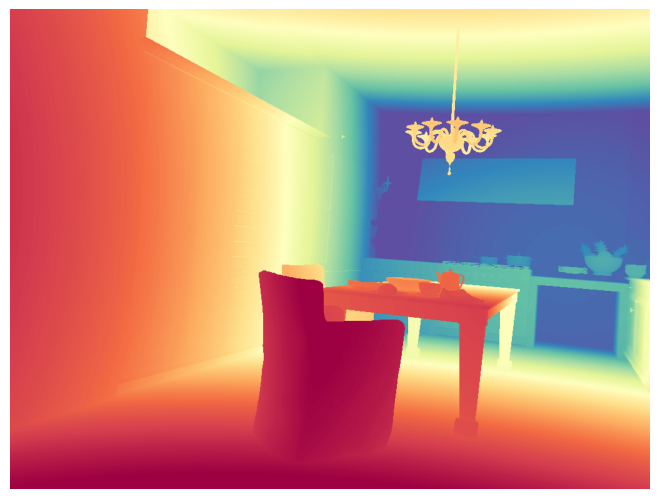}
        }
    \end{tabular}
    \vspace*{-0.2cm}
    \caption{Depth completion on Hypersim. \textit{Left}: Given partial depth. \textit{Middle}: Depth-estimate from the given partial depth. \textit{Right}: Ground-truth depth.}
    \label{fig:depthcompl}
\end{figure*}

\subsubsection{Training Details of the Depth Completion Model}
\label{supp:detals_depth_completion}

We finetune our model on synthetically generated partial depth maps, where only 2\% of pixels are available. In order to pass the sparse depth and its corresponding mask into the autoencoder, while losing as little information as possible, we first create a dense representation via nearest-neighbor interpolation and distance-functions respectively. More concretely, to represent the depth-mask for a pixel $p$, we store the value $\min_{p' \in \mathcal{M}} \|p' - p\|_2$, where $\mathcal{M}$ is the mask. 
We fine-tune our monocular depth estimation model first for 10k steps on Hypersim \cite{roberts2021hypersim} and then 8k more steps on NYU-v2 \cite{silberman12nyuv2} on a depth-completion task, where we further inflate the first convolutional layer to accept extra conditionings including sparse depth and sparse depth mask.
Given that the autoencoder is designed to handle non-sparse data, we upsample the sparse depth data using nearest neighbor interpolation and upsample the sparse depth mask using its distance function with $l_2$-norm.

\subsection{Additional Related Works}

\subsubsection{Monocular Depth Estimation}
The task of monocular depth estimation is usually approached in either a discriminative or a generative manner. For discriminative depth estimation, it can be categorized into regressing metric depth~\cite{SIDEreviewMERTAN2022103441, jun2022depth, bhat2021adabins,bhat2022localbins, yuan2022new, li2022binsformer, hu2024metric3d, yin2023metric3d} and relative depth~\cite{lee2019monocular, Ranftl2020MiDaS, Ranftl_2021_ICCV_DPT, SIDEreviewMERTAN2022103441, yang2024depthanything, yang2024depthanythingv2}.  

Various approaches have emerged within the realm of generative models~\cite{duan2023diffusiondepth,zhao2023vpd,saxena2023surprising_ddvm,ji2023ddp}, which aim to leverage diffusion models for metric depth estimation. For instance, DDP \cite{ji2023ddp} proposes an architecture to encode images but decode depth maps, achieving state-of-the-art results on the KITTI dataset. DepthGen \cite{saxena2023depthgen} extends a multi-task diffusion model to metric depth prediction which also handles the noisy ground truth. Its successor, DDVM \cite{saxena2023surprising_ddvm}, emphasizes pretraining on synthetic and real data for enhanced depth estimation capabilities. Please refer back to the main paper for generative approaches for relative depth estimations.

\subsubsection{Diffusion and Flow Matching}
Diffusion models~\cite{sohl2015deep,ho2020denoising,song2021scorebased} have excelled at various image generation tasks, including unconditional and class-conditional generation \cite{dhariwal2021diffusion,ho2022cascaded,sgdm}, image-to-image translation \cite{saharia2021image,sahariac-palette}, text-to-image synthesis \cite{rombach2022high,ramesh-dalle2,nichol-glide,sahariac-imagen}, and text-guided image editing \cite{instructpix2pix,ImagenEditor,hertz2022prompttoprompt,meng2021sdedit}. Additionally, they have been applied in discriminative tasks such as image enhancement~\cite{sahariac-palette}, panoptic segmentation~\cite{chen2022generalist}, and depth estimation~\cite{saxena2023depthgen}. On the other hand, flow-based methods like flow matching~\cite{lipman2022flow,rectifiedflow_iclr23,albergo2023stochastic} have gained considerable attention and have been explored in various domains, including image generation \cite{lipman2022flow,esser2024stablev3}, video prediction \cite{davtyan2022randomized}, human motion generation \cite{motionfm}, point cloud generation \cite{wu2022fast},  manifold data generation \cite{chen2023riemannian} and even text generation~\cite{HuEACL2024_flowseq}.

\subsubsection{Knowledge Distillation and Transfer Learning}
Knowledge Distillation, as introduced by \cite{hinton2015distilling}, is a method that takes one or multiple larger models trained for a specific task and transfers their knowledge to a smaller model. Training the smaller model on the label distribution predicted by the larger model(s), yields better performance than training on the actual ground truth labels.
Knowledge distillation is used in many contexts, from model compression \citep{sanh2020distilbert,aguilar2020knowledge,jin2019knowledge} to semi-supervised learning \citep{yang2024depthanything,tarvainen2017mean,xie2020self}, learning from noisy labels \citep{li2017learning}, or simply improving the performance of models via self-distillation \citep{zhang2019byot}. 

Transfer learning, on the other hand, focuses on fine-tuning a pre-trained model for a new but related task by leveraging the knowledge already embedded within the model. Most existing transfer learning methods attempt to align the feature representations across the two domains \citep{zhuang2020comprehensive,evci2022head2toe,zamir2018taskonomy}. Previous works have also explored transferring knowledge from heterogeneous teachers networks \citep{shen2019amalgamating,ye2019student}, i.e., multiple teachers trained on different tasks each, where the student learns to perform all tasks, or even completely cross-task setups \citep{ye2020distilling,auty2024learning}. In contrast, our generative depth estimation model utilizes prior knowledge derived from discriminative samples of the prior model.

\end{document}